\pdfoutput=1

\documentclass[11pt]{article}

\usepackage[final]{acl}

\usepackage{times}
\usepackage{latexsym}

\usepackage[T1]{fontenc}

\usepackage[utf8]{inputenc}

\usepackage{microtype}

\usepackage{inconsolata}

\usepackage{graphicx}

\usepackage{amsmath, amssymb}
\usepackage{booktabs}
\usepackage{lipsum}
\usepackage{multirow}
\usepackage{graphicx}
\usepackage{subcaption}
\usepackage{algorithm}
\usepackage{algorithmic}
\usepackage{afterpage}

\usepackage{xspace}
\usepackage[dvipsnames]{xcolor}

\newcommand{\ifprecedingtext}[1]{\ifvmode\relax\else#1\fi}

\newenvironment{redenv}{
    \color{BrickRed}
}{
    \ignorespacesafterend
}
\newcommand{\red}[1]{
    \begin{redenv}#1\end{redenv}  
}
\newenvironment{blueenv}{
    \color{blue}
}{
    \ignorespacesafterend
}

\newenvironment{orangeenv}{
    \color{orange}
}{
    \ignorespacesafterend
}

\newenvironment{purpleenv}{
    \color{black}
}{
    \ignorespacesafterend
}

\newenvironment{oliveenv}{
    \color{olive}
}{
    \ignorespacesafterend
}

\usepackage{makecell}
\usepackage[table]{xcolor} 
\usepackage{cleveref}
\newenvironment{greenenv}{
    \color{Green}
}{
    \ignorespacesafterend
}

\usepackage{svg} 
\usepackage{stfloats}
\usepackage{adjustbox}
\usepackage{xspace} 
\newcommand{\steam}[1][1.1em]{%
  \adjustbox{height=#1, raise=-.2\height}{\includegraphics{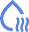}}
  \xspace
}
\usepackage{pifont}
\usepackage{threeparttable}
\newcommand{\cmark}{\textcolor{green!60!black}{\ding{51}}}
\newcommand{\xmark}{\textcolor{red!80!black}{\ding{55}}}
\newcommand{\ymark}{\textcolor{orange!80!black}{\textasciitilde}}

\title{Is Multilingual LLM Watermarking Truly Multilingual?\\ Scaling Robustness to 100+ Languages via Back-Translation}

\author{
  \textbf{Asim Mohamed\textsuperscript{1}},
  \textbf{Martin Gubri\textsuperscript{2,3}}
\\
\\
  \textsuperscript{1}African Institute for Mathematical Sciences,
  \textsuperscript{2}Parameter Lab,\\
  \textsuperscript{3}École Polytechnique, IP Paris, CNRS
\\
  \small{
    \textbf{Correspondence:} \href{mailto:amohamed@aimsammi.org}{amohamed@aimsammi.org}, \href{mailto:martin.gubri@parameterlab.de}{martin.gubri@parameterlab.de}
  }
}

\begin{document}

\maketitle
\begin{abstract}

Multilingual watermarking aims to make large language model (LLM) outputs traceable across languages, yet current methods still fall short. Despite claims of cross-lingual robustness, they are evaluated only on high-resource languages. We show that existing multilingual watermarking methods are not truly multilingual: they fail to remain robust under translation attacks in medium- and low-resource languages. We trace this failure to semantic clustering, which fails when the tokenizer vocabulary contains too few full-word tokens for a given language. To address this, we introduce STEAM, a detection method that uses Bayesian optimisation to search among 126 candidate languages for the back-translation that best recovers the watermark strength. It is compatible with any watermarking method, robust across different tokenizers and languages, non-invasive, and easily extendable to new languages. With average gains of +0.23 AUC and +37\%p TPR@1\%, STEAM provides a scalable approach toward fairer watermarking across the diversity of languages.

\end{abstract}

\section{Introduction}

Recent advances in multilingual watermarking claim to make large language model (LLM) outputs traceable across languages. Yet existing methods have been evaluated only on a small set of high-resource languages, leaving open the question of whether these techniques truly generalise to the world’s linguistic diversity. In this work, we show that \textit{current multilingual watermarking methods are not truly multilingual}. Their robustness weakens considerably for medium- and low-resource languages, revealing a major gap in current approaches to content provenance.

\begin{figure}[ht!] 
  \centering

  \begin{subfigure}{0.9\columnwidth}
    \centering
    \includegraphics[width=\linewidth]{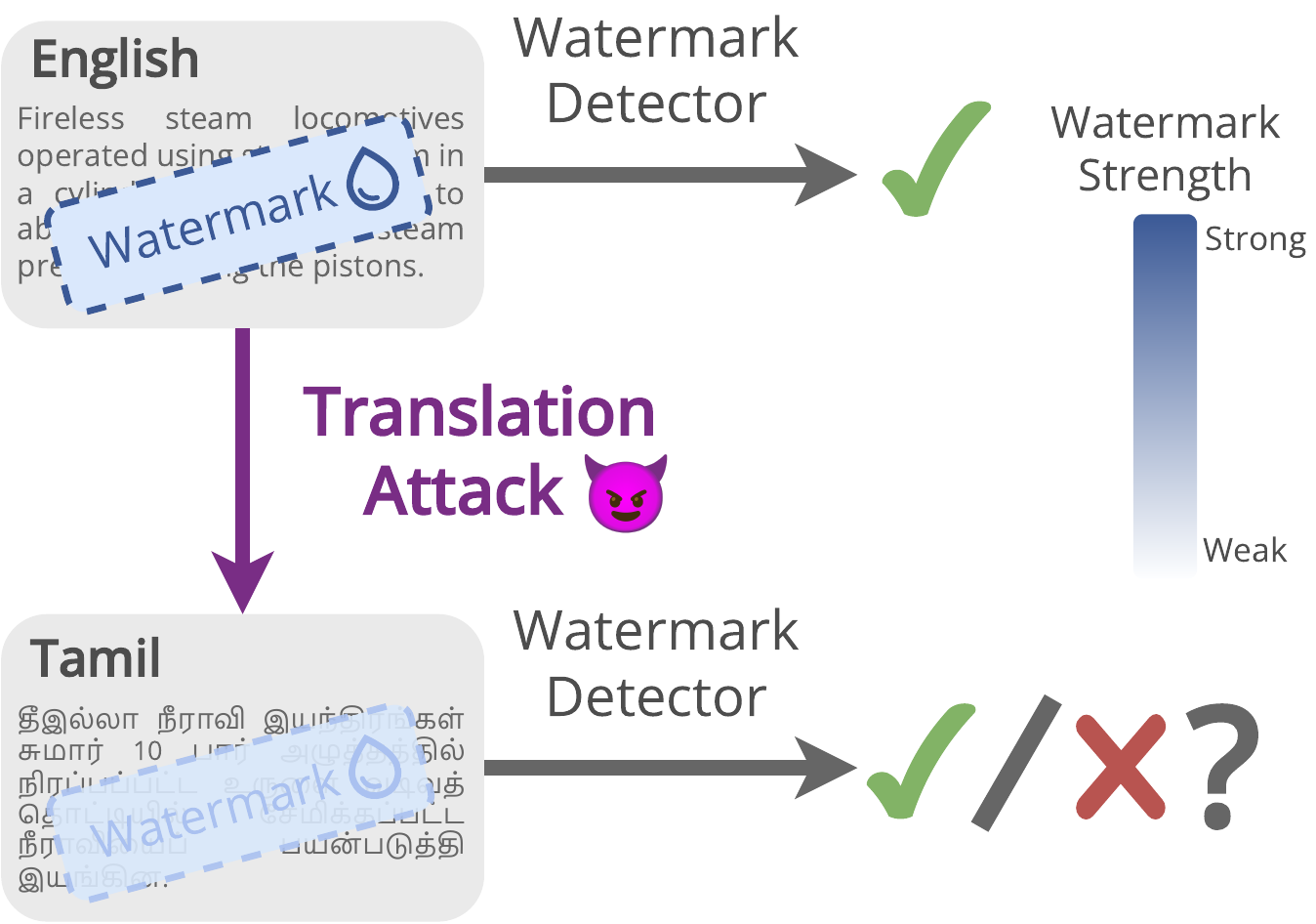}
    \caption{Translation attack against LLM watermarking.}
    \label{fig:teaser-translation-attack}
  \end{subfigure}

  \vspace{0.4em} 

  \begin{subfigure}{\columnwidth}
    \centering
    \includegraphics[width=0.9\linewidth]{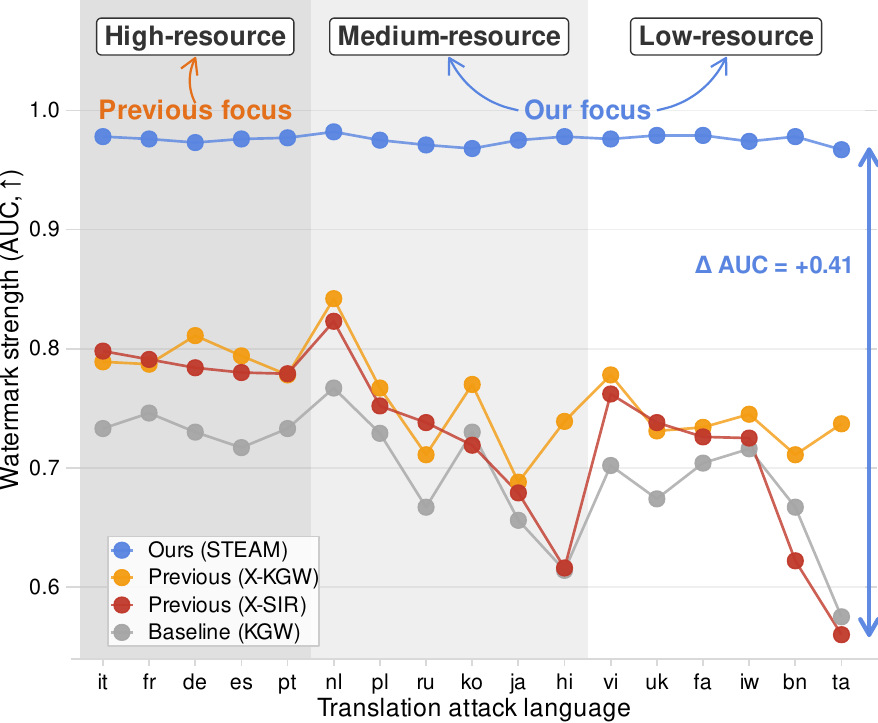}
    \caption{Watermarking robustness across languages.}
    \label{fig:teaser-auc-drop}
  \end{subfigure}

  \caption{(\subref{fig:teaser-translation-attack}) \textbf{Our goal} is to evaluate the robustness of LLM watermarks against translation attacks. (\subref{fig:teaser-auc-drop})~\textbf{Our analysis} reveals that existing multilingual watermarks fail to generalise across languages, while our approach (\textsc{Steam}\:\steam) performs consistently better across a wide range of languages overlooked by previous work.}
  \label{fig:teaser}
  \vspace{-1em}
\end{figure}

The limited robustness of multilingual watermarking has broad consequences. Watermarking was designed to identify LLM-generated text and to reduce the spread of misinformation on social media and synthetic content on the web. An adversary can exploit \textit{translation attacks}, in which a model generates text in one language and the content is translated into another, effectively scrubbing the watermark and reducing its strength \citep{he2024watermarkssurvivetranslationcrosslingual,ghanim2025uncovering,han2025synthid,luo2025lostoverlapexploringlogitbased,chen-etal-2025-improved}. Figure~\ref{fig:teaser-translation-attack} illustrates a translation attack. This threat is not theoretical: large-scale deployed systems such as Google's SynthID \citep{synthid2024}, used in Gemini, Veo, Imagen, and others, lose detectability after translation \citep{han2025synthid}. This vulnerability could enable undetectable synthetic content to spread in hundreds of languages, particularly in communities where moderation tools are less effective.

We focus specifically on translation attacks because they are the natural threat in multilingual deployment, where LLM-generated content frequently spreads across language communities via translation. Other attacks, such as paraphrasing or sample-based mixing, operate within the same language and preserve the overall token distribution. Several existing methods already target this regime through semantic invariance \citep{zhao2023provablerobustwatermarkingaigenerated,liu2024semanticinvariantrobustwatermark,ren2024robustsemanticsbasedwatermarklarge}. Translation attacks are different: they shift text to a different language, causing the token distribution to change substantially, which degrades the watermark signal in ways that semantic invariance cannot address.

\textit{Semantic clustering} has been proposed as a multilingual extension of watermarking. It groups semantically equivalent tokens (for example, `house', `maison', `casa') into clusters and treats all tokens in a cluster identically regarding the watermark key (for instance, all green or all red). While this approach performs adequately for high-resource languages, we observe that it performs poorly for many others. Tokenizers allocate tokens according to language frequency in their training data, meaning that only high-resource languages contain enough whole-word tokens to be properly represented in semantic clusters. For medium- and low-resource languages, most words are split into subword units not represented in any cluster, which substantially weakens the watermark. These findings suggest that semantic clustering cannot scale effectively beyond high-resource languages. 

To address the limited robustness of semantic clustering, we introduce \textbf{\textsc{Steam}}\:\steam (\textit{Search-based Translation-Enhanced Approach for Multilingual watermarking}), a detection-time method that uses Bayesian optimisation to select the back-translation language that best recovers watermark strength. \textsc{Steam} is non-invasive, model-agnostic, and compatible with any existing watermarking technique and tokenizer. \textsc{Steam} supports 126 candidate languages covering high-, medium-, and low-resource settings. Bayesian optimisation caps the number of evaluations at 20 per input, making the search tractable at this scale. The results show large and consistent performance gains over semantic clustering, with \textit{average improvements of +0.23 AUC and +37.4 percentage points (\textbf{\%p}) in TPR@1\%}. We perform an extensive robustness analysis and adaptive adversarial evaluation, all confirming \textsc{Steam}'s stability and effectiveness across diverse attack scenarios while maintaining a low false-positive rate.

Our contributions are:
\vspace{-0.5em}
\begin{enumerate}

    \item \textbf{Extensive multilingual evaluation.} We conduct a large-scale evaluation of multilingual watermarking methods across \mbox{high-,} \mbox{medium-,} and low-resource languages, uncovering weaknesses overlooked in prior work, which has focused exclusively on high-resource languages.

    \item \textbf{Analysis of the limitations of semantic clustering.} We identify that the limitations of current multilingual watermarking stem from their core reliance on clusters of tokens. 
    

    \item \textbf{\textsc{Steam}: a search-based, robust multilingual defence.} We introduce \textsc{Steam}\:\steam, which applies Bayesian optimisation to select the back-translation language that best recovers watermark strength. \textsc{Steam} already supports 126 candidate languages and is retroactively extensible to new ones, compatible with any watermarking technique and tokenizer, and non-invasive to the model output.

    \item \textbf{Robustness across diverse languages.} \textsc{Steam}\:\steam consistently outperforms existing multilingual watermarking methods, with improvements of up to \textbf{+0.41 AUC} and \textbf{+58.8\%p TPR@1\%}. 
    
\end{enumerate}

\section{Related Work}
LLM watermarking is generally classified by when the watermark is applied: training-time or inference-time (also known as logit-based) \cite{liu2024surveytextwatermarkingera}. We focus on the latter.

\paragraph{Logit-based watermarking.}
Logit-based watermarking embeds a watermark by directly modifying the token probability distribution (logits) during text generation \cite{liu2024surveytextwatermarkingera}. The seminal approach, KGW \cite{kirchenbauer23a}, partitions the tokenizer vocabulary into green and red lists using a random seed derived from a fixed window of previous tokens and biases generation towards green tokens. \citet{zhao2023provablerobustwatermarkingaigenerated} proposed Unigram Watermarking, an extension of KGW that employs a fixed green/red partition to improve robustness against text editing and paraphrasing attacks. 
To maintain text quality, \citet{hu2023unbiasedwatermarklargelanguage} introduced an unbiased watermarking approach that integrates watermarks without altering the overall probability distribution of the output. Several works \cite{lee2024wrotecodewatermarkingcode, lu2024entropybasedtextwatermarkingdetection, liu2024adaptivetextwatermarklarge,wu2024resilientaccessibledistributionpreservingwatermark} further improve robustness while preserving text quality.

Beyond these, ITS and EXP \cite{kuditipudi2024robustdistortionfreewatermarkslanguage} offer model-agnostic, distortion-free watermarking schemes that remain robust to text manipulation attacks. Our work analyses the multilingual capabilities of these techniques and builds upon them to develop our defence, \textsc{Steam} \steam.

\paragraph{Watermarking robustness.}
Several studies, including SIR \cite{liu2024semanticinvariantrobustwatermark}, SemaMark \cite{ren2024robustsemanticsbasedwatermarklarge}, semantic-aware watermarking \cite{fu2024watermarkingconditionaltextgeneration}, and SempStamp \cite{hou-etal-2024-semstamp}, incorporate semantic information to improve the robustness of watermarks against text transformation attacks. To achieve a balanced and context-aware partitioning of the green and red token lists, \citet{guo-etal-2024-context-aware} leveraged locality-sensitive hashing (LSH) \cite{10.1145/276698.276876} to generate a semantic key from contextual embeddings. Inspired by the inherent redundancy of multimedia data, WatME \cite{chen2024watmelosslesswatermarkinglexical} embeds mutual exclusion rules within the lexical space for text watermarking. Furthermore, \citet{luo2025lostoverlapexploringlogitbased} identified watermark collision, where multiple watermarks interact in ways that distort statistical distributions and hinder detection. These approaches primarily target attacks within a single language, such as paraphrasing or text editing, where the overall token distribution remains broadly stable. They do not address translation attacks, which shift text to a different language and cause a substantially different token distribution. This is a distinct failure mode that requires a different defence.

\paragraph{Multilingual watermarking.} 
While much of the initial research focused on monolingual English text, a growing body of work now addresses the unique challenges of cross-lingual watermarking. A foundational contribution in this area is X-SIR \cite{he2024watermarkssurvivetranslationcrosslingual}, a direct extension of the SIR framework designed to defend against translation attacks. Other works have focused on evaluating the cross-lingual robustness of existing methods. For example, \citet{han2025synthid} assessed the robustness of SynthID-Text \cite{synthid2024} to meaning-preserving transformations like back-translation. Similarly, \citet{ghanim2025uncovering} conduct a comparative evaluation of four watermarking methods: KGW, Unigram, EXP, and X-SIR. Their analysis assesses robustness and text quality under various parameters and removal attacks in cross-lingual settings. Although these studies provide valuable insights, their scope is often limited to high-resource languages. Our work addresses this gap by providing a more comprehensive cross-lingual evaluation that includes an extensive set of low- and medium-resource languages.

\begin{table}[t]
\centering
\setlength{\tabcolsep}{3pt}
\resizebox{\columnwidth}{!}{
\begin{tabular}{@{}lccc@{}}
\toprule
                   & KGW & X-KGW & STEAM \\
\textbf{Criterion} & \&\:SIR & \&\:X-SIR  & \steam(ours) \\
\midrule
\multicolumn{4}{@{}l}{\textit{Multilingual support}} \\[1pt]
\quad Multilingual & \xmark & \cmark & \cmark \\
\midrule
\multicolumn{4}{@{}l}{\textit{Multilingual robustness}} \\[1pt]
\quad Medium-resource    & \xmark & \ymark & \cmark \\
\quad Low-resource       & \xmark & \xmark & \;\cmark\textsuperscript{†} \\
\quad Tokenizer-robust & -- & \xmark & \cmark \\
\midrule
\multicolumn{4}{@{}l}{\textit{Multilingual integration}} \\[1pt]
\quad Non-invasive        & -- & \xmark & \cmark \\
\quad Watermark-agnostic  & -- & \xmark & \cmark \\
\quad Retroactive support & -- & \xmark & \cmark \\
\bottomrule
\end{tabular}
}
\caption{Comparison of watermarking methods and their multilingual capabilities. Criteria defined in App.~\ref{app:comparaison-def}.}
\label{tab:comparison}
\vspace{-0.1em}
\begin{tablenotes}
\footnotesize
\item \cmark~= Yes, \xmark~= No, \ymark~= Limited, --~= Not applicable
\item †~Requires translator (low-quality translation sufficient)
\end{tablenotes}
\vspace{-1em}
\end{table}

\section{Experimental Setup}
\label{sec:experimental_setup}
\paragraph{Dataset.} We base our evaluation on the English subset of the mC4 dataset \cite{raffel2023exploringlimitstransferlearning}, following the setup introduced by \citet{he2024watermarkssurvivetranslationcrosslingual}. We sample a test set of 500 texts for all experiments.

\paragraph{Attacks.} We evaluate watermark robustness under two translation-based attacks. Unless specified otherwise, all translations are performed with Google Translate. The first, direct translation, converts English outputs into a target language and is used in the main experiments. The second applies multi-step translation through a pivot language \citep{he2024watermarkssurvivetranslationcrosslingual} and is reported in Appendix~\ref{app:add-results}.

\paragraph{Multilingual models.} 
We use the following multilingual language models: \texttt{Aya-23-8B} \cite{aryabumi2024aya}, \texttt{LLaMA-3.2-1B} \cite{grattafiori2024llama3herdmodels}, and \texttt{LLaMAX-8B} \cite{luetal2024llamax}.

\paragraph{Watermarking methods.}
We analyse three watermarking schemes. We use the standard KGW \cite{kirchenbauer23a} as our primary non-multilingual baseline. Second, we evaluate X-SIR \cite{he2024watermarkssurvivetranslationcrosslingual}, a foundational work that proposes semantic clustering for cross-lingual robustness.  Finally, we introduce X-KGW, a method that applies semantic clustering to KGW. This setup allows us to isolate and measure the precise impact of semantic clustering on watermark robustness (see Appendix ~\ref{app:xkgw-description} for details about X-KGW).

\paragraph{Evaluation metrics.}  
We assess the strength of the watermark using two standard binary classification metrics, the Area Under the ROC Curve (\textit{AUC}) and the True Positive Rate at a false positive rate fixed at 1\% (\textit{TPR@1\%}).

\section{Semantic Clustering Fails in Diverse Multilingual Settings}
\label{sec:xsir_failure_modes}

We show that semantic clustering is not inherently multilingual: it lacks robustness in unsupported languages, and extending its coverage to more languages fails in medium- and low-resource settings. We then identify the structural cause of these failures.

\subsection{Robustness Against Unsupported Languages}
\label{subsec:unsupported_languages}

X-SIR is not robust when evaluated beyond its explicitly supported languages. The original evaluation covers only five supported languages \cite{he2024watermarkssurvivetranslationcrosslingual}, leaving robustness on unsupported ones unknown. Evaluating on ten unsupported languages (Italian, Spanish, Portuguese, Polish, Dutch, Croatian, Czech, Danish, Korean, and Arabic) shows that X-SIR is not robust in this setting (Appendix~\ref{app:unsupported_xsir}). X-SIR achieves average AUC and TPR@1\% of only 0.75 and 0.14 on Aya-23 8B, and 0.68 and 0.07 on LLaMA-3.2 1B. Some languages are more vulnerable than others: Arabic is the weakest for Aya-23 8B (AUC of 0.687, TPR@1\% of 0.093), while Portuguese and Arabic are the weakest for LLaMA-3.2 1B (AUC of 0.650, TPR@1\% of 0.055). Even a single poorly supported language is sufficient for an attacker to bypass detection. A hold-one-out experiment (Appendix~\ref{app:holdout-xsir}) confirms that the originally supported high-resource languages remain robust when one is withheld.

\subsection{Failure to Support a Broad Range of Languages}
\label{subsec:limited_tokens}

\begin{figure}
    \centering
    \includegraphics[width=\linewidth]{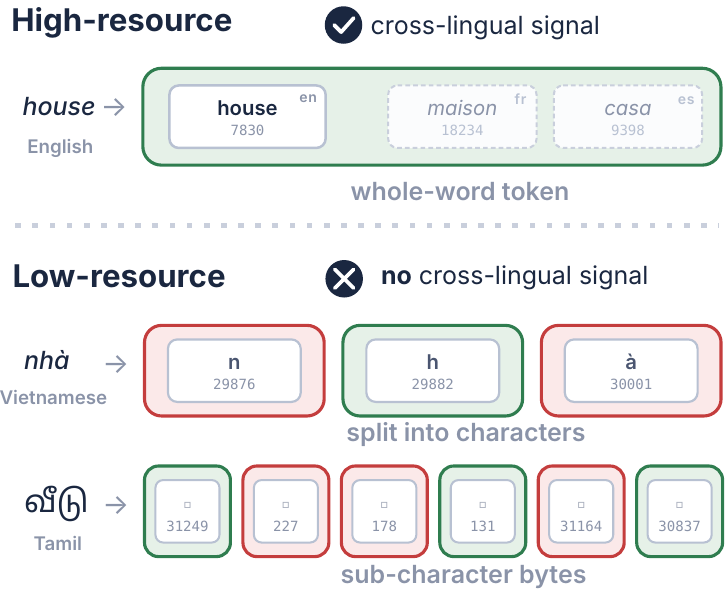}
    \caption{\textbf{Limitations of semantic clustering}. In high-resource languages (top), ``house'' and its translations (``maison'', ``casa'') are tokenised as whole words and join a single green-token cluster. In low-resource languages (bottom), ``house'' splits into character-level tokens (Vietnamese) or UTF-8 byte tokens (Tamil). Without a shared cluster, tokens carry no cross-lingual watermark signal (green \& red). LLaMA 3 tokeniser.}
    \label{fig:semantic-clust-limitattions}
    \vspace{-1em}
\end{figure}

\begin{table}[tb]
\centering
\small
\setlength{\tabcolsep}{1pt}
\begin{tabular}{cc|cccc}
\toprule
\multicolumn{2}{c}{\textbf{\makecell{Translation\\Attack}}} & \multicolumn{2}{c}{\textbf{X-SIR} ($\uparrow$)} & \multicolumn{2}{c}{\textbf{X-KGW} ($\uparrow$)} \\
\cmidrule(lr){1-2} \cmidrule(lr){3-4} \cmidrule(lr){5-6}
Type & Lang. & AUC & TPR@1\% & AUC & TPR@1\% \\
\midrule
\multirow{5}{*}{\makecell[c]{High-\\resource}}
 & fr & 0.791 & 0.149 & 0.787 & 0.280 \\
 & de & 0.784 & 0.163 & 0.811 & 0.312 \\
 & it & 0.798 & 0.152 & 0.789 & 0.354 \\
 & es & 0.780 & 0.150 & 0.794 & 0.278 \\
 & pt & 0.779 & 0.176 & 0.778 & 0.330 \\
\cmidrule[0.3pt](lr){1-6}
\multirow{6}{*}{\makecell[c]{Medium-\\resource}}
 & pl & 0.752 & 0.146 & 0.767 & 0.312 \\
 & nl & 0.823 & 0.213 & 0.842 & 0.332 \\
 & ru & 0.738 & 0.122 & 0.711 & 0.246 \\
 & hi & 0.616 & 0.056 & 0.739 & 0.194 \\
 & ko & 0.719 & 0.115 & 0.770 & 0.318 \\
 & ja & 0.679 & 0.103 & 0.688 & 0.160 \\
\cmidrule[0.3pt](lr){1-6}
\multirow{6}{*}{\makecell[c]{Low-\\resource}}
 & bn & 0.622 & 0.055 & 0.711 & 0.180 \\
 & fa & 0.726 & 0.131 & 0.734 & 0.242 \\
 & vi & 0.762 & 0.157 & 0.778 & 0.308 \\
 & iw & 0.725 & 0.115 & 0.745 & 0.220 \\
 & uk & 0.738 & 0.148 & 0.731 & 0.222 \\
 & ta & 0.560 & 0.049 & 0.737 & 0.172 \\
\cmidrule[0.3pt](lr){1-6}
\multicolumn{2}{r|}{Minimum} & 0.560 (ta) & 0.049 (ta) & 0.688 (ja) & 0.160 (ja) \\
\bottomrule
\end{tabular}
\caption{\textbf{Semantic clustering robustness tends to decrease from high- to low-resource languages, even with explicit support.} Aya-23 8B generates text in English; the translation attack targets each of 17 supported languages. Minimum is the worst-case robustness (best attack language). Other models in App.~\ref{app:supported_xsir}.}
\label{tab:xsir_auc_tpr01_translation_aya_grouped}
    \vspace{-1em}
\end{table}

Explicitly extending semantic clustering to more languages does not resolve its robustness failures.


We extend the support of X-SIR and X-KGW to 17 languages spanning high-, medium-, and low-resource settings (Appendix~\ref{app:language_categorization})%
\footnote{Our categorisation relies on the MUSE dictionary structure~\citep{conneau2017word}: high-resource languages possess extensive non-English-centric bidirectional mappings; medium- and low-resource languages are English-centric only, distinguished by dictionary size.}: French, German, Italian, Spanish, and Portuguese (high-resource); Polish, Dutch, Russian, Hindi, Korean, and Japanese (medium-resource); and Bengali, Persian, Vietnamese, Hebrew, Ukrainian, and Tamil (low-resource). Table~\ref{tab:xsir_auc_tpr01_translation_aya_grouped} reports results for Aya-23 8B. Robustness decreases with the level of resource: X-SIR's average AUC drops from 0.786 (high) to 0.721 (medium) to 0.689 (low), and X-KGW follows the same trend (0.792, 0.753, 0.739). The degradation is most severe for Tamil, where X-SIR reaches an AUC of only 0.560 and TPR@1\% of 0.049 on Aya-23 8B (0.561 on LLaMAX-3). Even with explicit support, semantic clustering fails to maintain reliable detection across all languages.
\subsection{On the Fundamental Limitations of Semantic Clustering in Multilingual Watermarking}
\label{subsec:semnantic_clustering_deadend}

The failures of X-SIR and X-KGW in mid- and low-resource languages stem from a structural property of semantic clustering: the method relies on full-word tokens, whose coverage in tokeniser vocabularies varies sharply across languages.

\begin{figure}[t]
    \centering
    \includegraphics[width=1.0\linewidth]{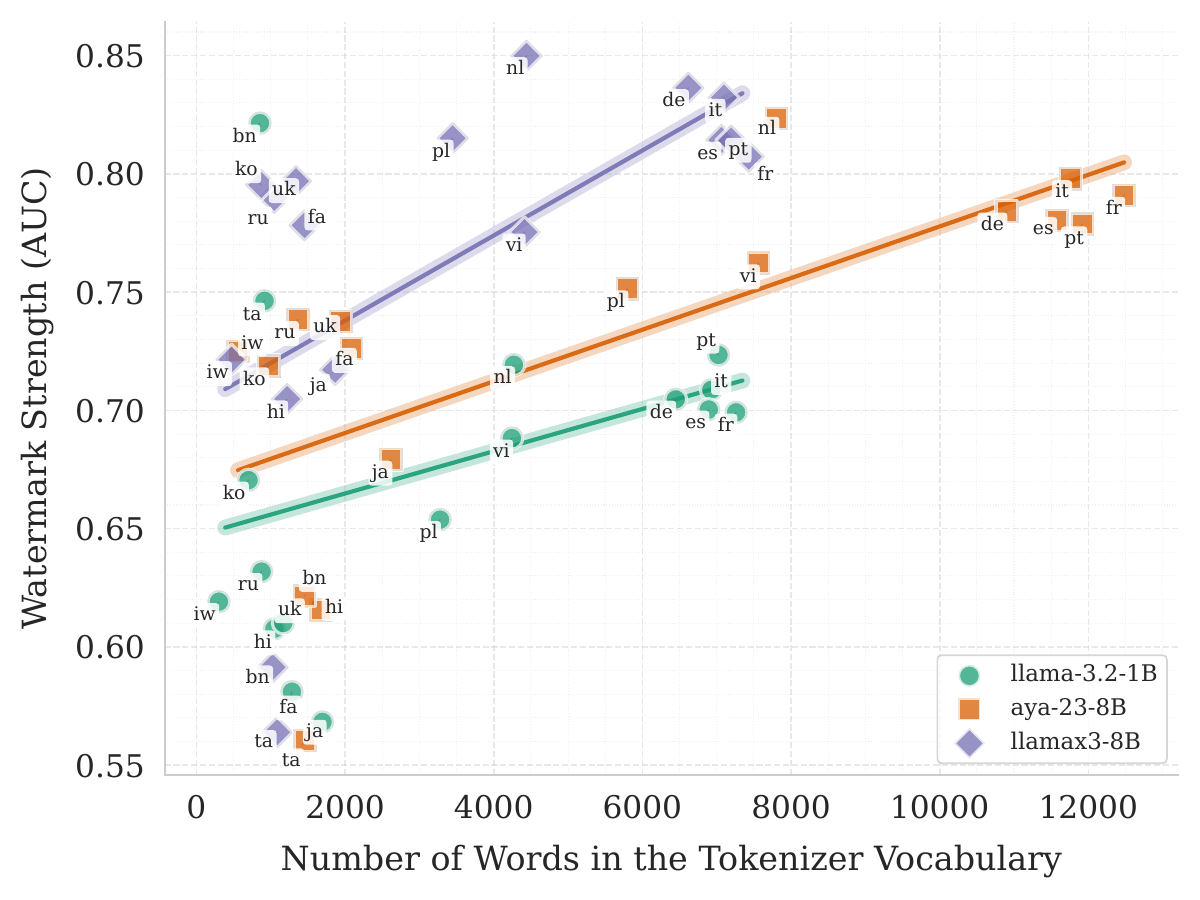}
    \caption{\textbf{Languages with larger tokenizer vocabularies have higher watermark robustness.} Average AUC per language and model across three seeds. Lines are least squared regressions.}
    \label{fig:tokenizer_words}
    \vspace{-1em}
\end{figure}

Semantic clustering assigns watermark signals using multilingual dictionaries to group semantically equivalent words across languages \cite{he2024watermarkssurvivetranslationcrosslingual}. However, the share of dictionary words that appear as whole-word tokens in tokeniser vocabularies varies sharply across languages (Appendix~\ref{app:tokenizer-dicts}). Low-resource languages have very few whole-word tokens, as low as 0.13\% for Hebrew. BPE-based tokenisers allocate tokens by frequency in training data, which favours high-resource languages and fragments others into subword units with limited to no semantic meaning (Figure~\ref{fig:semantic-clust-limitattions}).

\begin{figure*}
    \centering
    \includegraphics[width=\linewidth]{figures/steam_v4.pdf}
    \caption{\textbf{Overview of \textsc{Steam}} \steam. A suspect text (Bengali) is back-translated into candidate languages selected by Bayesian optimisation from a pool of 126 languages. Each language is represented by continuous syntactic and phonological features, which guide the search. A standard watermark detector computes a $z$-score for each candidate text. STEAM selects the back-translation language that yields the highest $z$-score, here English, the language of the watermarked text before the translation attack.}
    \label{fig:steam}
        \vspace{-1em}
\end{figure*}

Figure~\ref{fig:tokenizer_words} plots AUC against the number of whole-word tokens in the tokeniser vocabulary, for each language and model. A positive correlation emerges across all three models: languages with higher tokeniser coverage achieve stronger robustness, while those with lower coverage are far more vulnerable. This exposes three structural limitations. \textit{(i)} When a language has no whole-word tokens, X-KGW collapses to KGW, as no clusters can be formed (Figure~\ref{fig:semantic-clust-limitattions}). \textit{(ii)} Even multilingual tokenisers cannot resolve this, since BPE allocation inherently disadvantages underrepresented languages. \textit{(iii)} The vulnerability extends to monolingual watermarking: text generated in a high-resource language and translated into a low-resource one may lose its watermark entirely. These shortcomings are structural and cannot be overcome by expanding language support or retraining tokenisers.

\section{\textsc{Steam}: Scaling Defence to 100+ Languages via Back-Translation}
\label{sec:steam}

To address the limitations of semantic clustering for multilingual watermarking, we propose \textbf{Search-based Translation-Enhanced Approach for Multilingual watermarking (STEAM)} \steam, a novel, model-agnostic defence method that uses Bayesian optimisation to search for the back-translation language that best recovers watermark strength.

\subsection{STEAM Description}
\label{subsec:steam_mechanism}
\textsc{Steam}~\steam recovers a watermark strength degraded by translation attacks via multilingual back-translation. For each suspect text, \textsc{Steam} searches for the candidate language that, when used to back-translate the text, best recovers the watermark strength. Each candidate is evaluated using a watermark detector with a language-specific null hypothesis (described below). Finally, the highest corrected $z$-score across all evaluated candidates serves as the final test statistic. Figure~\ref{fig:steam} provides an overview of the \textsc{Steam} pipeline.

\paragraph{Bayesian optimisation for back-translation language selection.}
We use Bayesian Optimisation (BO) to search for the back-translation language that best recovers watermark strength. Each language is characterised by a 131-dimensional feature vector with syntactic and phonological properties sourced from URIEL~\cite{khan2025uriel}, a knowledge base of linguistic properties. BO first evaluates a small set of randomly selected candidates, then fits a Gaussian process surrogate that models the relationship between linguistic features and observed $z$-scores, allowing it to predict which unevaluated languages are likely to yield high $z$-scores based on their similarity to already evaluated ones. At each subsequent iteration, the next back-translation language is chosen by maximising the expected improvement over all unevaluated candidates. The process repeats until a fixed budget of 20 evaluations is exhausted, enabling \textsc{Steam} to scale to 126 candidate languages at low cost.

\paragraph{Language-specific null hypothesis.}
The standard $z$-score formula \citep{kirchenbauer23a} tests against a null hypothesis of a fixed green token fraction $\gamma$ (e.g. $\gamma = 50\%$), assumed uniform across all languages. This assumption is violated in low-resource languages, where tokenizers fragment single UTF-8 characters into sub-character tokens that are reused across many characters, making them disproportionately frequent. For instance, a single token accounts for 21.5\% of all tokens in our Tamil texts under Llama~3 tokenizer (see Appendix~\ref{app:subchar_dist}). If such a frequent token falls in the green list, the green token fraction exceeds $\gamma$ and the $z$-score inflates; when it falls in the red list, the $z$-score deflates. In both cases, the shift is independent of any watermark strength. Since \textsc{Steam} optimises for the highest $z$-score across candidate languages, such bias would cause it to systematically favour certain languages regardless of the watermark presence.

We address language token bias by replacing $\gamma$ with a language- and key-specific $\gamma_{\ell}$. Concretely, $\gamma_{\ell}$ is the empirical green token fraction measured on a calibration set of 500 human-written texts per candidate language, sourced from mC4 \cite{raffel2023exploringlimitstransferlearning}, for a fixed watermark key:

\begin{equation}
\label{eq:per-lang-gamma}
z_{\ell} = ( n_g - \gamma_{\ell}  n ) /  \sqrt{n \gamma_{\ell} (1 - \gamma_{\ell})}
\end{equation}

where $n_g$ is the number of green tokens of the suspect text and $n$ its total number of tokens. Ablating this correction causes only a modest AUC drop of 0.049, but correct language identification falls from 83.5\% to 38.6\% (App.~\ref{app:calibration-ablation}), confirming that normalisation is critical for unbiased detection.

\begin{table*}[tb]
\centering
\small
\setlength{\tabcolsep}{3pt}

\begin{tabular}{cc|cccc|cccc}
\toprule
 \multicolumn{2}{c}{\textbf{Source Language}}
 & \multicolumn{4}{c}{\textbf{AUC} ($\uparrow$)}
 & \multicolumn{4}{c}{\textbf{TPR@1\%} ($\uparrow$)} \\
 \cmidrule(lr){1-2} \cmidrule(lr){3-6}  \cmidrule(lr){7-10}
 Type & Language & KGW & X-KGW & X-SIR & STEAM\:\steam & KGW & X-KGW & X-SIR & STEAM\:\steam \\
\midrule
\multirow{5}{*}{\makecell[c]{High-\\resource}}
 & fr & 0.746 & 0.787 & 0.791 & \textbf{0.976} & 0.224 & 0.280 & \red{0.149} & \textbf{0.582} \\
 & de & 0.730 & 0.811 & 0.784 & \textbf{0.973} & 0.224 & 0.312 & \red{0.163} & \textbf{0.622} \\
 & it & 0.733 & 0.789 & 0.798 & \textbf{0.978} & 0.202 & 0.354 & \red{0.152} & \textbf{0.530} \\
 & es & 0.717 & 0.794 & 0.780 & \textbf{0.976} & 0.232 & 0.278 & \red{0.150} & \textbf{0.580} \\
 & pt & 0.733 & 0.778 & 0.779 & \textbf{0.977} & 0.246 & 0.330 & \red{0.176} & \textbf{0.538} \\
\cmidrule[0.3pt](lr){1-10}
\multirow{6}{*}{\makecell[c]{Medium-\\resource}}
 & pl & 0.729 & 0.767 & 0.752 & \textbf{0.975} & 0.228 & 0.312 & \red{0.146} & \textbf{0.526} \\
 & nl & 0.767 & 0.842 & 0.823 & \textbf{0.982} & 0.290 & 0.332 & \red{0.213} & \textbf{0.612} \\
 & ru & 0.667 & 0.711 & 0.738 & \textbf{0.971} & 0.158 & 0.246 & \red{0.122} & \textbf{0.576} \\
 & hi & 0.614 & 0.739 & 0.616 & \textbf{0.978} & 0.120 & 0.194 & \red{0.056} & \textbf{0.644} \\
 & ko & 0.730 & 0.770 & \red{0.719} & \textbf{0.968} & 0.210 & 0.318 & \red{0.115} & \textbf{0.460} \\
 & ja & 0.656 & 0.688 & 0.679 & \textbf{0.975} & 0.114 & 0.160 & \red{0.103} & \textbf{0.498} \\
\cmidrule[0.3pt](lr){1-10}
\multirow{6}{*}{\makecell[c]{Low-\\resource}}
 & bn & 0.667 & 0.711 & \red{0.622} & \textbf{0.978} & 0.068 & 0.180 & \red{0.055} & \textbf{0.604} \\
 & fa & 0.704 & 0.734 & 0.726 & \textbf{0.979} & 0.196 & 0.242 & \red{0.131} & \textbf{0.664} \\
 & vi & 0.702 & 0.778 & 0.762 & \textbf{0.976} & 0.186 & 0.308 & \red{0.157} & \textbf{0.626} \\
 & iw & 0.716 & 0.745 & 0.725 & \textbf{0.974} & 0.172 & 0.220 & \red{0.115} & \textbf{0.578} \\
 & uk & 0.674 & 0.731 & 0.738 & \textbf{0.979} & 0.210 & 0.222 & \red{0.148} & \textbf{0.542} \\
 & ta & 0.575 & 0.737 & \red{0.560} & \textbf{0.967} & 0.082 & 0.172 & \red{0.049} & \textbf{0.504} \\
\bottomrule
\end{tabular}
\caption{\textbf{STEAM \steam consistently and substantially outperforms semantic clustering.} Watermark strength (AUC and TPR@1\%) of multilingual watermarking techniques with 17 supported languages and Aya-23 8B. Red indicates robustness lower than the KGW baseline. Bolded is best. Other models and all supported languages in Appendix \ref{app:steam}.}
\label{tab:aya_translation_xsir_steam_grouped}
\vspace{-1em}
\end{table*}

\subsection{STEAM Evaluation}
\label{subsec:steam_evaluation}

    \paragraph{Comparison to semantic clustering.}

We evaluate \textsc{Steam} against semantic clustering methods to assess its robustness under translation attacks. As in \S\ref{subsec:limited_tokens}, all methods are tested on the same set of 17 attack languages. For back-translation, \textsc{Steam} chooses from a pool of 126 candidate languages covering a wide range of language families (Appendix \ref{app:bt-languages}). \textsc{Steam} achieves consistently strong results, maintaining an average AUC above \textbf{0.965} across all language categories, including medium- and low-resource ones (Table~\ref{tab:aya_translation_xsir_steam_grouped}, Appendix~\ref{app:steam}). Compared with semantic clustering approaches (X-SIR and X-KGW), \textsc{Steam} shows large gains: on average, +0.25 AUC and +44.0\%p TPR@1\% relative to X-SIR, and +0.216 AUC and +30.7\%p TPR@1\% relative to X-KGW. The largest improvements are observed for Tamil and Hindi, with up to +0.41 AUC and +58.8\%p TPR@1\%, respectively. These gains confirm that \textsc{Steam} generalises reliably beyond high-resource settings.

    \paragraph{STEAM robustness to unsupported languages.}

We next evaluate whether \textsc{Steam} remains robust when the best language is not in its pool of candidate languages. To ensure a fair comparison, we remove the language of the watermarked text (before the translation attack) from all methods: for semantic clustering, we exclude the corresponding dictionary; for \textsc{Steam}, we remove the language from the back-translation pool. In this setup, \textsc{Steam} maintains strong detection across all evaluated languages, with an average AUC of 0.967 (Table~\ref{tab:aya_translation_xsir_steam_unsupported_auc}). It consistently and substantially outperforms semantic clustering, with average gains of +0.19 AUC and +20.6\%p TPR@1\% over X-KGW and +0.22 AUC and +31.0\%p TPR@1\% over X-SIR. Unlike semantic clustering methods, which lose coverage when the language is absent from their dictionaries, \textsc{Steam} remains resilient: linguistically similar languages in its back-translation pool are sufficient to recover the watermark strength.

\begin{table}[tb]
\centering
\small
\setlength{\tabcolsep}{2pt}
\begin{tabular}{ccccccccc}
\toprule 
\multirow{2}{*}{\makecell[b]{\textbf{New}\\\textbf{Language}}} & \multicolumn{4}{c}{\textbf{AUC} ($\uparrow$)} \\
 \cmidrule(lr){2-5}
 & KGW & X-KGW & X-SIR & STEAM\:\steam \\
\midrule
it & 0.733 & 0.772 & 0.796 & \textbf{0.966} \\
es & 0.717 & 0.807 & 0.754 & \textbf{0.967} \\
pt & 0.732 & 0.792 & 0.775 & \textbf{0.971} \\
pl & 0.730 & 0.762 & 0.749 & \textbf{0.960} \\
nl & 0.768 & 0.808 & 0.776 & \textbf{0.966} \\
hr & 0.706 & 0.757 & 0.726 & \textbf{0.965} \\
cs & 0.717 & 0.754 & 0.773 & \textbf{0.974} \\
da & 0.713 & 0.764 & 0.734 & \textbf{0.971} \\
ko & 0.732 & 0.754 & \red{0.729} & \textbf{0.961} \\
ar & 0.689 & 0.765 & \red{0.687} & \textbf{0.971} \\
\bottomrule
\end{tabular}
\caption{\textbf{\textsc{Steam}\:\steam remains robust on unsupported languages and consistently outperforms multilingual methods.} Bolded is best. Red indicates lower than the KGW baseline. Full table in Appendix \ref{app:steam}.}
\label{tab:aya_translation_xsir_steam_unsupported_auc}
\vspace{-1em}
\end{table}

    \paragraph{Robustness across 126 languages.}

\begin{figure}[t]
    \centering
    \includegraphics[width=1.0\linewidth]{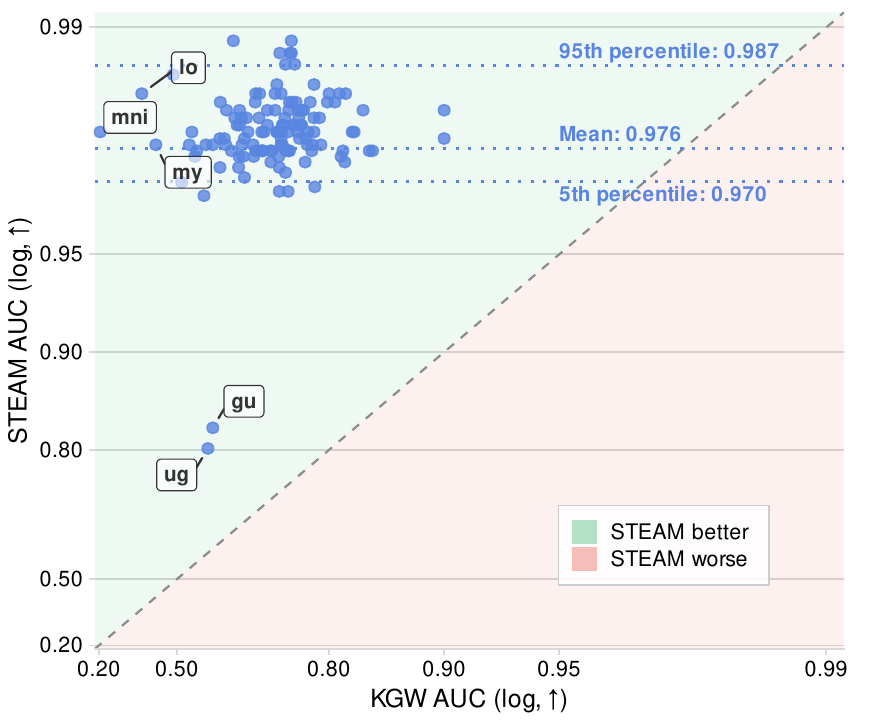}
    \caption{\textbf{\textsc{Steam} \steam generalises to 126 languages.} AUC of STEAM (y-axis) and KGW (x-axis) for 126 attack languages (Aya-23 8B). \textsc{Steam} always improves the detection of KGW-watermarked texts.}
    \label{fig:auc_steam_126languages}
    \vspace{-1em}
\end{figure}

We extend the evaluation to 126 attack languages, covering a broad range of language families and resource levels. Figure~\ref{fig:auc_steam_126languages} plots the AUC of \textsc{Steam} against KGW for each language. \textsc{Steam} achieves an average AUC of 0.976, and only two languages fall below 0.965. It improves detection over KGW on every attack language, confirming consistent robustness at scale.

    \paragraph{Robustness to the generation language.}

\textsc{Steam} remains robust when the watermarked text is not generated in English. Our other experiments watermark English text and attack it by translation out of English. We invert this setting, generating the watermarked text directly in each of the 17 languages and attacking it by translation into a language drawn at random from a pool of 40 languages. \textsc{Steam} achieves an average AUC of 0.971 and TPR@1\% of 0.504 across source languages (Appendix~\ref{app:source-language}), in line with the English-origin results.

\subsection{Robustness Analysis}
\label{subsec:steam_robustness}

    \paragraph{Robustness to translator mismatch.} 
The robustness of \textsc{Steam} should not depend on the specific translation service used. An adversary could try to bypass our defence by using a different translation system for their attack. 
We examine whether the performance of \textsc{Steam} remains robust when the attacker and defender use different translators. Table~\ref{tab:attacker_defender_pairs} reports the average AUC across German, Hindi, and Hebrew for all nine combinations of three translation services (Google Translate, DeepSeek-V3.2-Exp, and GPT-4o-mini). All pairs achieve an AUC above 0.94, and the diagonal (matched translator) does not consistently outperform mismatched settings. This suggests that our method genuinely recovers watermark strength independently of the specific translation service rather than relying on translator-specific artefacts. Detection is also possible locally with a small open-weights translator: using NLLB-200 distilled 600M keeps detection reliable, with a mean AUC of 0.907, though the TPR@1\% drops (App.~\ref{app:local_translator}).



    \paragraph{Adaptive evaluation: multistep translation attack.}

To assess the robustness of our defence under adaptive attack, we introduce a stronger multistep translation attack that adds an extra translation step beyond the single-hop setup. This design prevents \textsc{Steam} from relying on direct back-translation to recover the watermark strength. In this two-step attack, the text is first translated using the full set of languages from \S\ref{subsec:limited_tokens}, and the resulting output is then translated again through one of three pivot languages: German (high-resource), Korean (medium-resource), or Bengali (low-resource) (Appendix~\ref{app:steam}). Despite this adaptive setup, \textsc{Steam} remains robust, maintaining  an average AUC of 0.884 across all conditions. The lowest result occurs with a Korean pivot on high-resource languages (AUC of 0.833), while all other settings remain above 0.860. Although such multi-hop attacks can weaken other defences, they also tend to reduce overall translation quality, limiting their practical impact.

\paragraph{Additional robustness evaluation.}
We confirm \textsc{Steam}'s robustness across a broader range of scenarios. Under a combined paraphrase-then-translate attack, it retains AUC above 0.967 while X-SIR collapses to 0.492 (App.~\ref{app:paraphrase-then-translate-attacker}). Against a fully informed oracle adversary selecting the worst-case attack language, \textsc{Steam} still achieves AUC 0.802 (App.~\ref{app:oracle-attacker}). Detection takes approximately 17.85\,s per text, but is free with Google Translate and requires no GPU (App.~\ref{app:detection-cost}). A BO budget of 5 already achieves AUC 0.903, providing an option for faster detection (App.~\ref{app:bo-ablation}). Performance is stable across text lengths (App.~\ref{app:text-length-analysis}).

\begin{table}[tb]
\centering
\small
\setlength{\tabcolsep}{2pt}
\begin{tabular}{l ccc}
\toprule
\multirow{2}{*}{\textbf{Attacker}} & \multicolumn{3}{c}{\textbf{Defender}} \\
\cmidrule(l){2-4}
& G. Translate & DS-V3.2 & GPT-4o-mini \\
\midrule
Google Translate & 0.976 & 0.974 & 0.975 \\
DS-V3.2 & 0.979 & 0.952 & 0.943 \\
GPT-4o-mini & 0.976 & 0.973 & 0.964 \\
\bottomrule
\end{tabular}
\caption{\textbf{\textsc{Steam} \steam is robust to translator mismatch}. Average AUC (German, Hindi, Hebrew) for all attacker and defender translator combinations. Attacker translator is used for translation attack. Defender translator is used by \textsc{Steam}. DS-V3.2 is DeepSeek-V3.2-Exp.}
\label{tab:attacker_defender_pairs}
\vspace{-1em}
\end{table}


\subsection{False Positive Analysis}
    \label{sec:false-positive}

\paragraph{Calibrating the max-over-search statistic.}
Searching over candidate languages inflates the false-positive rate. \textsc{Steam} reports the highest corrected $z$-score among the languages it evaluates, so its null distribution is that of a maximum rather than of a single test. A naive single-test cutoff is therefore invalid: at $z=2.33$, the nominal 1\% threshold, the empirical FPR reaches $25.8\%$ for a pool of 126 languages (Table~\ref{tab:fpr-calibration}). We instead calibrate the decision threshold $\tau^*$ on the null distribution of the full max statistic. All results reported in this paper use this calibration.

\paragraph{Search budget governs the inflation.}
The inflation is driven by the number of evaluations the search performs, not by the size of the candidate pool. We run the full Bayesian optimisation search on held-out human-written texts, disjoint from the calibration texts used to estimate $\gamma_\ell$, and measure the empirical FPR averaged over 17 null languages. Under the fixed budget of 20 evaluations used throughout this paper, FPR is stable as the pool grows, from $29.2\%$ at $P=33$ to $25.8\%$ at $P=126$, and $\tau^*$ stays close to 3.9 (Table~\ref{tab:fpr-calibration}). Under a budget scaled with the pool ($30.3\%$ of $P$), FPR instead rises from $17.7\%$ to $41.2\%$. A fixed budget therefore keeps detection calibrated as language coverage grows.

\begin{table}[tb]
\centering
\small
\setlength{\tabcolsep}{3pt}
\begin{tabular}{lccc}
\toprule
\textbf{Budget} & \textbf{Pool Size} & \textbf{FPR ($z{=}2.33$) $(\downarrow)$} & \textbf{$\tau^*$} \\
\midrule
\multirow{3}{*}{Fixed}
 & 33  & 29.2\% & 3.94 \\
 & 66  & 25.8\% & 3.87 \\
 & 126 & 25.8\% & 3.90 \\
\midrule
\multirow{3}{*}{Proportional }
 & 33  & 17.7\% & 3.60 \\
 & 66  & 25.8\% & 3.87 \\
 & 126 & 41.2\% & 4.07 \\
\bottomrule
\end{tabular}
\caption{\textbf{Search budget governs \textsc{Steam}'s false-positive rate}. Empirical FPR of the max-over-search statistic on held-out human texts. We report FPR at the naive single-test 1\% cutoff ($z=2.33$) and at $\tau^*$, a threshold calibrated to the 99th percentile of the null max-score distribution. A \textit{fixed} budget evaluates 20 candidate languages whatever the pool size, as done throughout this paper, while a \textit{proportional} budget evaluates $30.3\%$ of the pool.}
\label{tab:fpr-calibration}
\end{table}

\paragraph{Why the inflation stays slow.}
Two properties of the corrected scores bound how fast the null maximum grows. First, the language-specific null hypothesis (Eq.~\ref{eq:per-lang-gamma}) removes the per-language bias that would otherwise let some languages win the maximum regardless of watermark presence (\S\ref{subsec:steam_mechanism}). Second, the corrected null scores are approximately standard Gaussian, and by extreme-value theory the maximum of $P$ such scores concentrates around $\sqrt{2\ln P}$, which grows only from $\approx 2.6$ at $P=33$ to $\approx 3.1$ at $P=126$. This bound is conservative, since typologically related languages yield positively correlated scores and thus fewer effective independent tests.

\paragraph{Pool size trades detection power for coverage.}
Enlarging the candidate pool costs detection power, but far too slowly to erase \textsc{Steam}'s advantage. We measure TPR@1\% for pools of 33, 66 and 126 languages, sampled at random from the full set (Table~\ref{tab:pool-size-sensitivity}). TPR@1\% generally decreases as the pool grows, since a fixed budget of 20 evaluations then covers a smaller fraction of the candidates. Yet at $P=126$, \textsc{Steam} still outperforms X-KGW and X-SIR on every language.

\begin{table}[tb]
\centering
\small
\setlength{\tabcolsep}{3pt}
\begin{tabular}{l cccccc}
\toprule
& \multicolumn{6}{c}{\textbf{TPR@1\%} ($\uparrow$)} \\
\cmidrule(lr){2-7}
Lang. & KGW & X-KGW & X-SIR & \multicolumn{3}{c}{STEAM \steam} \\
\cmidrule(lr){5-7}
 & & & & (P=33) & (P=66) & (P=126) \\
\midrule
fr & 0.224 & 0.280 & \red{0.149} & \textbf{0.830} & 0.696 & 0.582 \\
de & 0.224 & 0.312 & \red{0.163} & \textbf{0.792} & 0.710 & 0.622 \\
it & 0.202 & 0.354 & \red{0.152} & \textbf{0.710} & 0.612 & 0.530 \\
es & 0.232 & 0.278 & \red{0.150} & \textbf{0.758} & 0.622 & 0.580 \\
hi & 0.120 & 0.194 & \red{0.056} & \textbf{0.808} & 0.580 & 0.644 \\
ko & 0.210 & 0.318 & \red{0.115} & \textbf{0.708} & 0.700 & 0.460 \\
ja & 0.114 & 0.160 & \red{0.103} & \textbf{0.622} & 0.508 & 0.498 \\
bn & 0.068 & 0.180 & \red{0.055} & 0.464 & \textbf{0.634} & 0.604 \\
\bottomrule
\end{tabular}
\caption{\textbf{\textsc{Steam} \steam remains above all baselines across all pool sizes.} TPR@1\% for KGW, X-KGW, X-SIR, and \textsc{Steam} \steam with back-translation pools of size 33, 66, and 126. Supporting more languages requires a larger pool, which trades increased robustness in new languages for a slight increase in the false positive rate. Even at P=126, \textsc{Steam} \steam consistently outperforms all baselines across all languages. Red indicates values fall below the undefended KGW baseline. Bolded is best.}
\label{tab:pool-size-sensitivity}
\end{table}
\section{Conclusion}

Current multilingual watermarking methods fail to remain robust under translation attacks, especially in medium- and low-resource languages. If watermarking lacks robustness in a given language, online content in that language may be disproportionately affected by synthetic or undesirable content. This risk is especially serious for low- and medium-resource languages, which already face a shortage of high-quality digital resources and often lack effective moderation systems.

To address this, we introduced \textsc{Steam} \steam, a watermark-agnostic method that applies Bayesian optimisation to select the back-translation language that best recovers watermark strength. Extensive experiments scaling to 126 languages and diverse attack scenarios show that \textsc{Steam} achieves consistently stronger robustness and fairness than existing multilingual watermarking methods, particularly in medium- and low-resource settings.

Our findings highlight the need for watermarking research to treat linguistic diversity as a core requirement, ensuring that the security and trust of large language models extend to all languages, not only those with abundant digital resources.

\clearpage

\section*{Limitations}

\textsc{Steam} is designed for translation attacks only. Multilingual robustness failures stem from structural properties of tokeniser vocabularies, which underrepresent low- and medium-resource languages (\S\ref{subsec:semnantic_clustering_deadend}). This specificity has motivated dedicated multilingual watermarking techniques, confirming that multilinguality constitutes a distinct research challenge. Other threat models, such as paraphrasing, are addressed by separate methods \citep{zhao2023provablerobustwatermarkingaigenerated,liu2024semanticinvariantrobustwatermark,ren2024robustsemanticsbasedwatermarklarge}. A key property of translation attacks is their invertibility: \textsc{Steam} recovers watermark strength by translating back into a language close to the original. Paraphrasing offers no such inverse operation, which means \textsc{Steam} addresses translation attacks only.

\textsc{Steam} requires access to a translation service for the target language. Such services may not exist for the most under-resourced languages. However, LLMs themselves tend to perform poorly in languages without translation support, so the practical scope of the limitation is narrow. Additionally, \textsc{Steam} does not require high-quality translation: our robustness experiments show that different translation services yield comparable results (\S\ref{subsec:steam_robustness}).

\textsc{Steam} involves a trade-off between detection cost and detection accuracy. Each suspect text requires several back-translations, which introduce latency and, when a paid translation service is used, a monetary cost. This overhead affects detection only, not generation: users experience no added latency, since \textsc{Steam} does not modify the generation. The search budget controls both the cost and the accuracy of detection, so a smaller budget yields cheaper but less accurate detection (Appendices~\ref{app:bo-ablation},~\ref{app:detection-cost}). A comparable trade-off applies to language coverage, where a larger candidate pool extends protection to more attack languages but reduces detection power (\S\ref{sec:false-positive}). \textsc{Steam} is therefore appropriate for offline detection, such as provenance auditing or post-hoc moderation, but not for real-time or streaming verification.

Finally, despite extensive testing, the implementation of our experiments may contain implementation errors.

\section*{Ethical Considerations}

This work has potential dual-use implications. On one hand, studying adversarial attacks against watermarking could inform malicious actors about possible strategies to weaken watermark defences. However, we believe the benefits outweigh these risks. 

First, our contribution is not only an analysis but also a concrete defence (\textsc{Steam}) that achieves a high level of robustness, substantially exceeding prior multilingual watermarking methods. Our results demonstrate that \textsc{Steam} provides consistently strong robustness against translation attacks across a wide range of languages. 

Second, by explicitly addressing low- and medium-resource languages, our method promotes fairness: watermarking becomes more reliable across diverse linguistic settings, rather than being limited to a handful of high-resource languages.

Robust multilingual watermarking is an important safeguard against misuse of large language models, such as the generation and dissemination of fake news or disinformation in less-resourced languages where moderation tools are often weaker. We view this work as a step toward improving the security and trustworthiness of multilingual AI systems.

\textsc{Steam} translates the suspect text at detection time. Using a third-party translation API therefore exposes that text to an external provider, which is undesirable when detection is applied to sensitive material. \textsc{Steam} makes no assumption about the translator, so detection can run entirely locally with an open-weights translation model or a self-hosted LLM, and no text needs to leave the detector's infrastructure. Our translator mismatch results confirm that no specific service is required (\S\ref{subsec:steam_robustness}), and we further find that a small open-weights translation model run locally is sufficient for reliable detection. We recommend local translation whenever the text under examination is sensitive.


\bibliography{acl_latex}

\appendix
\section*{\Large Appendix}

The appendices contain the following sections:

\begin{itemize}
    \item Appendix \ref{app:xp-setting} details the experimental settings,
    \item Appendix \ref{app:add-results} contains additional experimental results of Section~\ref{sec:xsir_failure_modes} about semantic clustering,
    \item Appendix \ref{app:steam}  contains additional experimental results of Section~\ref{sec:steam} about Steam,
    \item Appendix \ref{app:ai-assist} contains our usage of AI assistants.
    \item Appendix \ref{app:artifacts} contains the details of our artifacts.
\end{itemize}

For transparency and reproducibility, our code is available on GitHub at \url{https://github.com/asimzz/steam}

\section{Experimental Setting}
\label{app:xp-setting}

    \subsection{Hyperparameters}
To ensure reproducibility, we detail the hyperparameters used for both the neural network training and the watermark generation/detection phases of our experiments.

    \paragraph{X-SIR neural network training.} The neural network component of X-SIR, which inherits its architecture from SIR, was trained using the following hyperparameters:
\begin{itemize}
    \item Architecture. The model consists of 4 layers, with an input dimension of 1024, a hidden dimension of 500, and an output dimension of 300.
    \item Optimization. We used \textit{Stochastic Gradient Descent (SGD)} with a \textit{learning rate} of 0.006 and a \textit{weight decay} of 0.2. A \texttt{StepLR} scheduler with a \textit{step size} of 200 and a gamma of 0.1 was employed to adjust the learning rate during training.
    \item Training. The model was trained for 2000 epochs with a batch size of 32.
\end{itemize}

    \paragraph{Watermarking scheme parameters.}
    For the watermark generation and detection phases, the following parameters were used for each scheme:
    \begin{itemize}
        \item KGW. We used the default parameters recommended by \citet{kirchenbauer23a}: a green list proportion (gamma) of 0.25, a logit bias (delta) of 2.0, and the \texttt{minhash} seeding scheme.
        \item X-KGW. To create a direct comparison with XSIR, we set the context width to 1. The gamma and delta values were kept consistent with KGW at 0.25 and 2.0, respectively.
        \item X-SIR. We followed the original implementation, setting the window size to 5, the chunk size to 10, and the logit bias (delta) to 1.0. The multilingual sentence embeddings were generated using the \texttt{paraphrase-multilingual-mpnet-base-v2} model.
    \end{itemize}

    \subsection{Computational Resources \& Softwares}

All experiments were conducted on a Google Cloud Platform instance of type \texttt{n1-standard-4}, equipped with 4 vCPUs, 15 GB of RAM, and two NVIDIA T4 GPUs.

All translations use the Google Translate service accessed through the \href{https://github.com/nidhaloff/deep-translator}{\texttt{deep\_translator}} Python library, which provides a unified interface to various translation APIs. The translator mismatch experiment in \S \ref{subsec:steam_robustness} employs the DeepSeek API for back-translation.

We used Pytorch as our deep learning framework \cite{paszke2019pytorchimperativestylehighperformance}, with CUDA support for GPU acceleration. In addition, we employed Hugging Face Transformers library\footnote{\url{https://huggingface.co/}} \cite{wolf-etal-2020-transformers} to access pretrained models and tokenizers.

\subsection{Back-Translation Candidate Languages}
\label{app:bt-languages}

STEAM searches over a pool of candidate back-translation languages using Google Translate as the default translation service.
Table~\ref{tab:bt-languages} lists all candidate languages used in our experiments for \S~\ref{subsec:steam_evaluation} \S~\ref{subsec:steam_robustness}.
These languages span diverse language families, scripts, and resource levels, covering
high-resource (e.g., English, French, German), medium-resource (e.g., Hungarian, Romanian, Thai),
and low-resource (e.g., Tigrinya, Xhosa, Yoruba) settings.

\begin{table*}[h!]                                                                                                                                                                             
  \centering                                       
  \small                                                                                                                                                                                         
  \begin{tabular}{ll|ll|ll}                                                                                                                                                                      
  \toprule                                                                                                                                                                                       
  \textbf{Code} & \textbf{Language} & \textbf{Code} & \textbf{Language} & \textbf{Code} & \textbf{Language} \\                                                                                                                                         
  \midrule                                                                                                                                                                                       
  af & Afrikaans         & iw & Hebrew              & fa & Persian          \\                                                                                                                   
  sq & Albanian          & hi & Hindi               & pl & Polish           \\                                                                                                                   
  am & Amharic           & hu & Hungarian           & pt & Portuguese       \\                                                                                                                   
  ar & Arabic            & is & Icelandic           & pa & Punjabi          \\                                                                                                                   
  hy & Armenian          & ig & Igbo                & ro & Romanian         \\                                                                                                                   
  as & Assamese          & ilo & Ilocano            & ru & Russian          \\
  ay & Aymara            & id & Indonesian          & sm & Samoan           \\                                                                                                                   
  az & Azerbaijani       & ga & Irish               & sa & Sanskrit         \\                                                                                                                   
  bm & Bambara           & it & Italian             & gd & Scottish Gaelic  \\                                                                                                                   
  eu & Basque            & ja & Japanese            & nso & Sepedi          \\                                                                                                                   
  be & Belarusian        & jw & Javanese            & sr & Serbian          \\                                                                                                                 
  bn & Bengali           & kn & Kannada             & st & Sesotho          \\                                                                                                                   
  bho & Bhojpuri         & kk & Kazakh              & sn & Shona            \\                                                                                                                 
  bs & Bosnian           & km & Khmer               & sd & Sindhi           \\                                                                                                                   
  bg & Bulgarian         & rw & Kinyarwanda         & si & Sinhala          \\                                                                                                                 
  ca & Catalan           & ko & Korean              & sk & Slovak           \\                                                                                                                   
  ceb & Cebuano          & kri & Krio               & sl & Slovenian        \\                                                                                                                 
  ny & Chichewa          & ku & Kurdish (Kurmanji)  & so & Somali           \\                                                                                                                   
  zh-CN & Chinese (Simp.)& ckb & Kurdish (Sorani)   & es & Spanish          \\                                                                                                                 
  zh-TW & Chinese (Trad.)& ky & Kyrgyz              & su & Sundanese        \\                                                                                                                   
  co & Corsican          & lo & Lao                 & sw & Swahili          \\                                                                                                                 
  hr & Croatian          & la & Latin               & sv & Swedish          \\                                                                                                                   
  cs & Czech             & lv & Latvian             & tg & Tajik            \\                                                                                                                   
  da & Danish            & ln & Lingala             & ta & Tamil            \\                                                                                                                   
  dv & Dhivehi           & lt & Lithuanian          & tt & Tatar            \\                                                                                                                   
  nl & Dutch             & lg & Luganda             & te & Telugu           \\                                                                                                                 
  en & English           & lb & Luxembourgish       & th & Thai             \\                                                                                                                   
  eo & Esperanto         & mk & Macedonian          & ti & Tigrinya         \\                                                                                                                 
  et & Estonian          & mai & Maithili           & ts & Tsonga           \\                                                                                                                   
  ee & Ewe               & mg & Malagasy            & tr & Turkish          \\                                                                                                                 
  tl & Filipino          & ms & Malay               & tk & Turkmen          \\                                                                                                                   
  fi & Finnish           & ml & Malayalam           & ak & Twi              \\                                                                                                                 
  fr & French            & mt & Maltese             & uk & Ukrainian        \\                                                                                                                   
  fy & Frisian           & mi & Maori               & ur & Urdu             \\                                                                                                                 
  gl & Galician          & mr & Marathi             & ug & Uyghur           \\                                                                                                                   
  ka & Georgian          & mni-Mtei & Meitei        & uz & Uzbek            \\                                                                                                                   
  de & German            & lus & Mizo               & vi & Vietnamese       \\                                                                                                                   
  el & Greek             & mn & Mongolian           & cy & Welsh            \\                                                                                                                   
  gu & Gujarati          & my & Myanmar             & xh & Xhosa            \\                                                                                                                 
  ht & Haitian Creole    & ne & Nepali              & yi & Yiddish          \\                                                                                                                   
  ha & Hausa             & no & Norwegian           & yo & Yoruba           \\                                                                                                                 
  haw & Hawaiian         & or & Odia                & zu & Zulu             \\                                                                                                                   
  \bottomrule                                                                                                                                                                                  
  \end{tabular}                                                                                                                                                                                  
  \caption{All 126 candidate back-translation languages supported by \textsc{Steam} \steam.}                                                                                                                     
  \label{tab:bt-languages}                                                                                                                                                                       
  \end{table*}

    \subsection{Description of X-KGW}
    \label{app:xkgw-description}
X-KGW (Cross-lingual KGW) is a hybrid watermarking approach we introduce to combine the hash-based mechanism of KGW with the semantic clustering strategy of X-SIR. Unlike KGW, which partitions individual tokens, X-KGW operates at the cluster level. The process consists of three distinct phases:
\begin{enumerate}
    \item Semantic cluster construction. Following the X-SIR framework, we first construct a multilingual semantic graph using bilingual translation dictionaries. The Louvain community detection algorithm is then applied to partition the vocabulary $\mathcal{V}$ into $\mathcal{C}$ disjoint semantic clusters, yielding a token-to-cluster mapping $m:\mathcal{V} \rightarrow {0,1,\dots,\mathcal{C}-1}$.
    \item Hash-based cluster partitioning. During text generation, at each timestep $t$, a context window of preceding tokens $(w_{t-h}, \dots, w_{t-1})$ is used to compute a hash-based seed. This seed is then employed to pseudo-randomly partition the $\mathcal{C}$ clusters into green and red sets, with a fraction $\gamma$ designated as green.
    \item Cluster-based logit modification. Finally, a positive bias $\delta$ is applied to the logits of all tokens belonging to clusters assigned to the green set. The model then samples the next token from this modified probability distribution.
\end{enumerate}

By combining KGW logit biasing with semantic clustering, X-KGW seeks to preserve watermark robustness under multilingual transformations while maintaining detection accuracy.

    \subsection{Multilingual Dictionaries \& Language Categorization}
    \label{app:language_categorization}

    To construct our multilingual dictionary, we relied on the MUSE dictionary \cite{conneau2017word}, the same resource used by \citet{he2024watermarkssurvivetranslationcrosslingual} to build the semantic clusters. In addition to its role in dictionary-based clustering, we used MUSE to categorize the 17 languages included in our evaluation of \S \ref{subsec:limited_tokens}, \S \ref{subsec:semnantic_clustering_deadend}, \S \ref{subsec:steam_evaluation}, \S \ref{subsec:steam_robustness}.

    \begin{itemize}
        \item A language was marked \textit{high-resource} if it possesses extensive, non-English-centric dictionary mappings (i.e., bidirectional dictionaries with multiple other languages in the set).
        \item In contrast, languages whose resources are primarily English-centric, where MUSE provides only bidirectional dictionaries with English, were classified as either \textit{medium-resource} or \textit{low-resource}. The distinction between these two groups was determined by the size (i.e., the number of word pairs) of their respective English dictionaries.
    \end{itemize}

    \subsection{Definitions of Multilingual Watermarking Comparison Criteria}
    \label{app:comparaison-def}

For clarity, we provide the definitions of the criteria used in Table~\ref{tab:comparison} to compare multilingual watermarking techniques:

\begin{itemize}
    \item \textbf{Multilingual.} Designed to resist translation attacks.
    \item \textbf{Medium/low-resource.} Robust against translation attacks into medium- or low-resource languages.
    \item \textbf{Tokenizer-robust.} Robustness against translation attacks does not depend on the tokenizer.
    \item \textbf{Non-invasive.} Multilingual support does not modify generation logits, so text quality is preserved.
    \item \textbf{Watermark-agnostic.} Compatible with any watermarking technique without modification.
    \item \textbf{Retroactive support.} New languages can be added without regenerating the watermark key; already-watermarked texts remain detectable.
\end{itemize}

    \subsection{Prompt for DeepSeek-V3.2-Exp Translation}
    To use DeepSeek-V3.2-Exp as a translation engine, we designed a structured prompt format. We define:

{\small\begin{verbatim}
Source language: {src_lang}
Target language: {tgt_lang}
Input text: {response}
\end{verbatim}}

    \texttt{src\_lang} and \texttt{tgt\_lang} indicate the source and target language codes. We convert these codes into their full language names using the \texttt{Language} class from the \texttt{langcodes} \footnote{\url{https://pypi.org/project/langcodes/}} library:
{\small\begin{verbatim}
Language.make(language=src_lang).display_name()
Language.make(language=tgt_lang).display_name()
\end{verbatim}}

    The final prompt provided to DeepSeek-V3.2-Exp:

{\small\begin{verbatim}
Translate the following
{Language.make(language=src_lang).display_name()}
text to
{Language.make(language=tgt_lang).display_name()}:

{response}
\end{verbatim}}

    \subsection{Bayesian Optimisation Details}
\label{app:bo_details}

\paragraph{Surrogate model.}
We use a single-task Gaussian Process (GP) with a constant mean function as implemented by \texttt{SingleTaskGP} in BoTorch~\citep{balandat2020botorch}. The GP hyperparameters (kernel lengthscales, output scale, and noise variance) are optimised by maximising the exact marginal log-likelihood using GPyTorch~\citep{gardner2021gpytorch}. The GP is refitted from scratch at each BO iteration.

\paragraph{Acquisition function.}
We use Log Expected Improvement~(\texttt{LogExpectedImprovement} in BoTorch), a numerically stable variant of expected improvement that operates in log-space to avoid vanishing gradients when the current best value is far from the predictive mean. Since the search space is a finite set of discrete languages rather than a continuous domain, we evaluate the acquisition function at all unevaluated candidate feature vectors and select the one with the highest value.

\paragraph{Language feature vectors.}
Each candidate probe language is represented by a 131-dimensional feature vector obtained by concatenating \texttt{syntax\_knn} (103 dimensions) and \texttt{phonology\_knn} (28 dimensions) from the URIEL typological database~\citep{khan2025uriel}, accessed via \texttt{lang2vec}~\citep{littell-etal-2017-uriel}. Feature vectors are pre-computed once and reused across all texts.

\paragraph{Budget.}
Each text is allocated a maximum of 20 evaluations (3 initial + 17 BO iterations).

\paragraph{Implementation.}
The full pipeline is implemented in Python using BoTorch 0.11+, GPyTorch 1.11+, and PyTorch 2.0+.

\vspace{5em}

\section{Additional Results About Semantic Clustering}
\label{app:add-results}

This Appendix contains the additional results of Section \ref{sec:xsir_failure_modes}.

    \subsection{Hold-out languages for X-SIR}
    \label{app:holdout-xsir}

    Tables~\ref{tab:xsir_auc_tpr01_combined_before_after_aya} and~\ref{tab:xsir_auc_tpr01_combined_before_after_llama} report hold-one-out results for Aya-23~8B and LLaMA-3.2~1B respectively.

    \begin{table*}[h!]
\centering
\footnotesize
\setlength{\tabcolsep}{2pt}
\begin{tabular}{cc|cccccc}
\toprule
\multicolumn{2}{c}{\textbf{Languages}} & \multicolumn{3}{c}{\textbf{AUC} ($\uparrow$)} & \multicolumn{3}{c}{\textbf{TPR@1\%} ($\uparrow$)} \\
\cmidrule(lr){1-2} \cmidrule(lr){3-5} \cmidrule(lr){6-8}
Held-out & Prompt & Held-Out & Supported & $\Delta$ & Held-Out & Supported & $\Delta$ \\
\midrule
\multirow{3}{*}{en}
 & fr & 0.795 {\tiny±0.045} & 0.816 {\tiny±0.014} & +0.021 & 0.198 {\tiny±0.049} & 0.149 {\tiny±0.042} & \red{-0.049} \\
 & de & 0.780 {\tiny±0.054} & 0.811 {\tiny±0.018} & +0.031 & 0.172 {\tiny±0.047} & 0.168 {\tiny±0.039} & \red{-0.004} \\
 & zh & 0.731 {\tiny±0.020} & 0.669 {\tiny±0.042} & \red{-0.062} & 0.141 {\tiny±0.011} & 0.083 {\tiny±0.014} & \red{-0.058} \\
\cmidrule[0.3pt](lr){1-8}
\multirow{3}{*}{fr}
 & en & 0.757 {\tiny±0.022} & 0.799 {\tiny±0.027} & +0.042 & 0.157 {\tiny±0.016} & 0.139 {\tiny±0.037} & \red{-0.018} \\
 & de & 0.723 {\tiny±0.020} & 0.781 {\tiny±0.025} & +0.058 & 0.101 {\tiny±0.029} & 0.156 {\tiny±0.061} & +0.055 \\
 & zh & 0.651 {\tiny±0.026} & 0.638 {\tiny±0.037} & \red{-0.013} & 0.076 {\tiny±0.021} & 0.052 {\tiny±0.020} & \red{-0.024} \\
\cmidrule[0.3pt](lr){1-8}
\multirow{3}{*}{de}
 & en & 0.736 {\tiny±0.011} & 0.802 {\tiny±0.020} & +0.067 & 0.153 {\tiny±0.050} & 0.214 {\tiny±0.035} & +0.061 \\
 & fr & 0.765 {\tiny±0.004} & 0.784 {\tiny±0.020} & +0.019 & 0.139 {\tiny±0.068} & 0.118 {\tiny±0.039} & \red{-0.021} \\
 & zh & 0.667 {\tiny±0.041} & 0.642 {\tiny±0.014} & \red{-0.025} & 0.073 {\tiny±0.032} & 0.065 {\tiny±0.016} & \red{-0.008} \\
\cmidrule[0.3pt](lr){1-8}
\multirow{3}{*}{zh}
 & en & 0.644 {\tiny±0.052} & 0.692 {\tiny±0.041} & +0.048 & 0.120 {\tiny±0.046} & 0.111 {\tiny±0.011} & \red{-0.009} \\
 & fr & 0.671 {\tiny±0.072} & 0.714 {\tiny±0.045} & +0.043 & 0.112 {\tiny±0.053} & 0.069 {\tiny±0.025} & \red{-0.043} \\
 & de & 0.675 {\tiny±0.048} & 0.701 {\tiny±0.026} & +0.026 & 0.105 {\tiny±0.053} & 0.107 {\tiny±0.027} & +0.002 \\
\cmidrule[0.3pt](lr){1-8}
\multirow{3}{*}{ja}
 & en & 0.685 {\tiny±0.059} & 0.656 {\tiny±0.017} & \red{-0.029} & 0.113 {\tiny±0.024} & 0.070 {\tiny±0.008} & \red{-0.043} \\
 & fr & 0.698 {\tiny±0.038} & 0.670 {\tiny±0.018} & \red{-0.028} & 0.101 {\tiny±0.033} & 0.089 {\tiny±0.023} & \red{-0.012} \\
 & de & 0.688 {\tiny±0.037} & 0.669 {\tiny±0.027} & \red{-0.019} & 0.138 {\tiny±0.031} & 0.079 {\tiny±0.005} & \red{-0.059} \\
& zh & 0.681 {\tiny±0.043} & 0.658 {\tiny±0.003} & \red{-0.023} & 0.110 {\tiny±0.046} & 0.093 {\tiny±0.016} & \red{-0.017} \\
\bottomrule
\end{tabular}
\caption{\textbf{Hold-one-out: X-SIR is slightly sensitive to the set of supported languages.} $\Delta$ = Supported $-$ Held-out, where \textit{Supported} uses all five original languages (en, fr, de, zh, ja) and \textit{Held-out} excludes the target language from the cluster; the attack always targets the held-out language. Aya-23~8B generates text in the Prompt language. Differences are small ($|\Delta| \leq 0.07$ for AUC) but variable across configurations and models, indicating mild sensitivity to which languages are included. Red indicates negative $\Delta$.}
\label{tab:xsir_auc_tpr01_combined_before_after_aya}
\end{table*}

    \begin{table*}[h!]
\centering
\footnotesize
\setlength{\tabcolsep}{2pt}
\begin{tabular}{cc|cccccc}
\toprule
\multicolumn{2}{c}{\textbf{Languages}} & \multicolumn{3}{c}{\textbf{AUC} ($\uparrow$)} & \multicolumn{3}{c}{\textbf{TPR@1\%} ($\uparrow$)} \\
\cmidrule(lr){1-2} \cmidrule(lr){3-5} \cmidrule(lr){6-8}
Held-out & Prompt & Held-Out & Supported & $\Delta$ & Held-Out & Supported & $\Delta$ \\
\cmidrule[0.3pt](lr){1-8}
\multirow{3}{*}{en}
 & fr & 0.901{\tiny±0.017} & 0.907{\tiny±0.015} & +0.006 & 0.363{\tiny±0.047} & 0.275{\tiny±0.007} & \red{-0.088} \\
 & de & 0.884{\tiny±0.043} & 0.894{\tiny±0.041} & +0.010 & 0.379{\tiny±0.131} & 0.331{\tiny±0.043} & \red{-0.048} \\
 & zh & 0.795{\tiny±0.043} & 0.827{\tiny±0.013} & +0.032 & 0.450{\tiny±0.040} & 0.407{\tiny±0.011} & \red{-0.043} \\
\cmidrule[0.3pt](lr){1-8}
\multirow{3}{*}{fr}
 & en & 0.743{\tiny±0.050} & 0.682{\tiny±0.026} & \red{-0.061} & 0.113{\tiny±0.056} & 0.068{\tiny±0.012} & \red{-0.045} \\
 & de & 0.787{\tiny±0.050} & 0.687{\tiny±0.011} & \red{-0.100} & 0.161{\tiny±0.038} & 0.093{\tiny±0.007} & \red{-0.068} \\
 & zh & 0.678{\tiny±0.006} & 0.681{\tiny±0.007} & +0.003 & 0.086{\tiny±0.027} & 0.083{\tiny±0.005} & \red{-0.003} \\
\cmidrule[0.3pt](lr){1-8}
\multirow{3}{*}{de}
 & en & 0.693{\tiny±0.017} & 0.692{\tiny±0.043} & \red{-0.001} & 0.069{\tiny±0.022} & 0.068{\tiny±0.025} & \red{-0.001} \\
 & fr & 0.724{\tiny±0.023} & 0.725{\tiny±0.044} & +0.001 & 0.098{\tiny±0.011} & 0.071{\tiny±0.009} & \red{-0.027} \\
 & zh & 0.665{\tiny±0.018} & 0.693{\tiny±0.046} & +0.028 & 0.065{\tiny±0.009} & 0.073{\tiny±0.014} & +0.008 \\
\cmidrule[0.3pt](lr){1-8}
\multirow{3}{*}{zh}
 & en & 0.666{\tiny±0.035} & 0.605{\tiny±0.012} & \red{-0.061} & 0.082{\tiny±0.052} & 0.026{\tiny±0.006} & \red{-0.056} \\
 & fr & 0.704{\tiny±0.011} & 0.609{\tiny±0.024} & \red{-0.095} & 0.085{\tiny±0.041} & 0.036{\tiny±0.012} & \red{-0.049} \\
 & de & 0.703{\tiny±0.022} & 0.636{\tiny±0.017} & \red{-0.067} & 0.088{\tiny±0.010} & 0.050{\tiny±0.008} & \red{-0.038} \\
\cmidrule[0.3pt](lr){1-8}
\multirow{4}{*}{ja}
 & en & 0.576{\tiny±0.052} & 0.573{\tiny±0.024} & \red{-0.003} & 0.039{\tiny±0.019} & 0.033{\tiny±0.005} & \red{-0.006} \\
 & fr & 0.630{\tiny±0.066} & 0.581{\tiny±0.041} & \red{-0.049} & 0.067{\tiny±0.016} & 0.025{\tiny±0.014} & \red{-0.042} \\
 & de & 0.624{\tiny±0.055} & 0.589{\tiny±0.039} & \red{-0.035} & 0.074{\tiny±0.023} & 0.037{\tiny±0.018} & \red{-0.037} \\
 & zh & 0.663{\tiny±0.059} & 0.650{\tiny±0.026} & \red{-0.013} & 0.121{\tiny±0.065} & 0.075{\tiny±0.040} & \red{-0.046} \\
\bottomrule
\end{tabular}
\caption{\textbf{Hold-one-out: X-SIR is slightly sensitive to the set of supported languages.} $\Delta$ = Supported $-$ Held-out, where \textit{Supported} uses all five original languages (en, fr, de, zh, ja) and \textit{Held-out} excludes the target language from the cluster; the attack always targets the held-out language. LLaMA-3.2~1B generates text in the Prompt language. Differences are small ($|\Delta| \leq 0.10$ for AUC) but variable across configurations and models, indicating mild sensitivity to which languages are included. Red indicates negative $\Delta$.}
\label{tab:xsir_auc_tpr01_combined_before_after_llama}
\end{table*}
    
    \clearpage
    \subsection{Unsupported languages for X-SIR \& X-KGW}
    \label{app:unsupported_xsir}

    \begin{table}[H]
\centering
\small
\setlength{\tabcolsep}{3pt}
\begin{tabular}{ccccc}
\toprule
 \multirow{2}{*}{\makecell[b]{\textbf{New}\\\textbf{Lang.}}}
 & \multicolumn{2}{c}{\textbf{X-SIR} ($\uparrow$)} & \multicolumn{2}{c}{\textbf{X-KGW} ($\uparrow$)} \\
\cmidrule(lr){2-3} \cmidrule(lr){4-5}
 & AUC & TPR@1\% & AUC & TPR@1\% \\
\midrule
it & 0.796 & 0.177 & 0.772 & 0.238 \\
es & 0.754 & 0.155 & 0.807 & 0.230 \\
pt & 0.775 & 0.133 & 0.792 & 0.286 \\
pl & 0.749 & 0.127 & 0.762 & 0.236 \\
nl & 0.776 & 0.164 & 0.808 & 0.314 \\
hr & 0.726 & 0.124 & 0.757 & 0.210 \\
cs & 0.773 & 0.111 & 0.754 & 0.254 \\
da & 0.734 & 0.161 & 0.764 & 0.266 \\
ko & 0.729 & 0.136 & 0.754 & 0.226 \\
ar & 0.687 & 0.093 & 0.765 & 0.168 \\
\cmidrule[0.3pt](lr){1-5}
Min.  & 0.687 (ar) & 0.093 (ar) & 0.754 (cs, ko) & 0.168 (ar) \\
\bottomrule
\end{tabular}
\caption{\textbf{Semantic clustering is weak for unsupported languages}. Watermark strength (AUC and TPR@1\%) of X-SIR and X-KGW, limited to the five originally supported languages (en, fr, de, zh, ja). Aya-23 8B generates English text, which is then translated into a new unsupported language for evaluation. Minimum marks the weakest robustness (best attack case).} 
\label{tab:xsir_xkgw_auc_tpr01_translation_vs_translation_aya}
\vspace{-1em}
\end{table}
    
\begin{table}[H]
\centering
\small
\setlength{\tabcolsep}{3pt}
\begin{tabular}{ccccc}
\toprule
 \multirow{2}{*}{\makecell[b]{\textbf{New}\\\textbf{Lang.}}}
 & \multicolumn{2}{c}{\textbf{X-SIR} ($\uparrow$)} & \multicolumn{2}{c}{\textbf{X-KGW} ($\uparrow$)} \\
\cmidrule(lr){2-3} \cmidrule(lr){4-5}
 & AUC & TPR@1\% & AUC & TPR@1\% \\
\midrule
it & 0.699 & 0.069 & 0.760 & 0.212 \\
es & 0.665 & 0.076 & 0.744 & 0.222 \\
pt & 0.641 & 0.059 & 0.722 & 0.152 \\
pl & 0.679 & 0.069 & 0.677 & 0.144 \\
nl & 0.754 & 0.095 & 0.781 & 0.244 \\
hr & 0.660 & 0.066 & 0.733 & 0.162 \\
cs & 0.650 & 0.064 & 0.759 & 0.190 \\
da & 0.675 & 0.093 & 0.765 & 0.196 \\
ko & 0.673 & 0.062 & 0.672 & 0.124 \\
ar & 0.655 & 0.055 & 0.704 & 0.168 \\
\cmidrule[0.3pt](lr){1-5}
Min. & 0.641 (pt) & 0.055 (ar) & 0.672 (ko) & 0.124 (ko) \\
\bottomrule
\end{tabular}
\caption{\textbf{Semantic clustering is weak for unsupported languages}. Watermark strength (AUC and TPR@1\%) of X-SIR and X-KGW, limited to the five originally supported languages (en, fr, de, zh, ja). LLaMA-3.2 1B generates English text, which is then translated into a new unsupported language for evaluation. Minimum marks the weakest robustness (best attack case).}

\label{tab:xsir_xkgw_auc_tpr01_translation_llama3_unsupported}
\end{table}

\begin{table}[H]
\centering
\small
\setlength{\tabcolsep}{3pt}
\begin{tabular}{ccccc}
\toprule
 \multirow{2}{*}{\makecell[b]{\textbf{New}\\\textbf{Lang.}}}
 & \multicolumn{2}{c}{\textbf{X-SIR} ($\uparrow$)} & \multicolumn{2}{c}{\textbf{X-KGW} ($\uparrow$)} \\
\cmidrule(lr){2-3} \cmidrule(lr){4-5}
 & AUC & TPR@1\% & AUC & TPR@1\% \\
\midrule
it & 0.829 & 0.335 & 0.860 & 0.510 \\
es & 0.810 & 0.314 & 0.864 & 0.490 \\
pt & 0.812 & 0.337 & 0.861 & 0.498 \\
pl & 0.812 & 0.308 & 0.817 & 0.420 \\
nl & 0.845 & 0.351 & 0.896 & 0.584 \\
hr & 0.804 & 0.279 & 0.822 & 0.344 \\
cs & 0.798 & 0.305 & 0.838 & 0.444 \\
da & 0.832 & 0.357 & 0.871 & 0.436 \\
ko & 0.792 & 0.276 & 0.800 & 0.390 \\
ar & 0.765 & 0.251 & 0.815 & 0.358 \\
\cmidrule[0.3pt](lr){1-5}
Min. & 0.765 (ar) & 0.251 (ar) & 0.800 (ko) & 0.344 (hr) \\
\bottomrule
\end{tabular}
\caption{\textbf{Semantic clustering is weak for unsupported languages}. Watermark strength (AUC and TPR@1\%) of X-SIR and X-KGW, limited to the five originally supported languages (en, fr, de, zh, ja). LLaMAX-3 8B generates English text, which is then translated into a new unsupported language for evaluation. Minimum marks the weakest robustness (best attack case).}
\label{tab:xsir_xkgw_auc_tpr01_translation_llamax_unsupported}
\end{table}

\medskip

\begin{table*}[H]
\centering
\small
\setlength{\tabcolsep}{6pt}
\centering
\begin{tabular}{ccccccc}
\toprule
 \textbf{CWRA Attack} & \multicolumn{2}{c}{\textbf{Aya-23 8B} ($\uparrow$)} & \multicolumn{2}{c}{\textbf{LLaMA-3.2 1B} ($\uparrow$)} & \multicolumn{2}{c}{\textbf{LLaMAX-3 8B} ($\uparrow$)} \\
\cmidrule(lr){1-1} \cmidrule(lr){2-3} \cmidrule(lr){4-5} \cmidrule(lr){6-7}
\multicolumn{1}{c|}{\textbf{New Language}} & AUC & TPR@1\% & AUC & TPR@1\% & AUC & TPR@1\% \\
\midrule
it & 0.746 & 0.194 & 0.855 & 0.245 & 0.826 & 0.313 \\
es & 0.751 & 0.147 & 0.830 & 0.217 & 0.816 & 0.319 \\
pt & 0.781 & 0.179 & 0.854 & 0.244 & 0.827 & 0.336 \\
pl & 0.793 & 0.195 & 0.859 & 0.289 & 0.815 & 0.299 \\
nl & 0.836 & 0.252 & 0.900 & 0.375 & 0.835 & 0.360 \\
hr & 0.810 & 0.236 & 0.853 & 0.269 & 0.790 & 0.291 \\
cs & 0.785 & 0.180 & 0.835 & 0.201 & 0.787 & 0.309 \\
da & 0.857 & 0.247 & 0.864 & 0.243 & 0.830 & 0.331 \\
ko & 0.750 & 0.157 & 0.852 & 0.255 & 0.809 & 0.309 \\
ar & 0.704 & 0.209 & 0.822 & 0.222 & 0.771 & 0.286 \\
\cmidrule[0.3pt](lr){1-7}
Minimum & 0.704 (ar) & 0.147 (es) & 0.822 (ar) & 0.201 (cs) & 0.771 (ar) & 0.286 (ar)\\
\bottomrule
\end{tabular}
\label{tab:xsir_auc_tpr01_cwra_aya_llamax_llama3_unsupported}
\caption{\textbf{Semantic clustering (XSIR) performs inconsistently on an expanded set of supported languages}. The semantic clustering is applied using an expanded set of 17 newly supported languages. A prompt in English is first translated into each target language. Aya-23 8B, LLaMA-3.2 1B, and LLaMAX-3 8B are then prompted with the translated input to generate text in the target language. Finally, the CWRA attack is applied by translating the generated text back into English. Baseline is the average on the original supported languages. Higher values indicate better robustness. Minimum indicates the worst-case robustness, i.e., the best language for an attack.}
\end{table*}

    \clearpage
    \subsection{Supported languages for X-SIR \& X-KGW}
    \label{app:supported_xsir}

    \begin{table}[ht]
\centering
\small
\setlength{\tabcolsep}{1pt}
\begin{tabular}{cc|cccc}
\toprule
\multicolumn{2}{c}{\textbf{Translation Attack}} & \multicolumn{2}{c}{\textbf{X-SIR} ($\uparrow$)} & \multicolumn{2}{c}{\textbf{X-KGW} ($\uparrow$)} \\
\cmidrule(lr){1-2} \cmidrule(lr){3-4} \cmidrule(lr){5-6}
Type & Language & AUC & TPR@1\% & AUC & TPR@1\% \\
\midrule
\multirow{5}{*}{\makecell[c]{High-\\resource}}
 & fr & 0.702 & 0.085 & 0.719 & 0.166 \\
 & de & 0.708 & 0.067 & 0.752 & 0.186 \\
 & it & 0.712 & 0.111 & 0.750 & 0.230 \\
 & es & 0.703 & 0.089 & 0.724 & 0.222 \\
 & pt & 0.726 & 0.102 & 0.747 & 0.206 \\
\cmidrule[0.3pt](lr){1-6}
\multirow{6}{*}{\makecell[c]{Medium-\\resource}}
 & pl & 0.657 & 0.065 & 0.703 & 0.188 \\
 & nl & 0.722 & 0.091 & 0.787 & 0.252 \\
 & ru & 0.635 & 0.075 & 0.656 & 0.100 \\
 & hi & 0.611 & 0.037 & 0.620 & 0.084 \\
 & ko & 0.673 & 0.055 & 0.701 & 0.154 \\
 & ja & 0.571 & 0.042 & 0.598 & 0.110 \\
\cmidrule[0.3pt](lr){1-6}
\multirow{6}{*}{\makecell[c]{Low-\\resource}}
 & bn & 0.825 & 0.509 & 0.701 & 0.078 \\
 & fa & 0.584 & 0.055 & 0.673 & 0.086 \\
 & vi & 0.691 & 0.084 & 0.722 & 0.186 \\
 & iw & 0.622 & 0.026 & 0.702 & 0.134 \\
 & uk & 0.613 & 0.064 & 0.725 & 0.118 \\
 & ta & 0.749 & 0.095 & 0.672 & 0.108 \\
\cmidrule[0.3pt](lr){1-6}
\multicolumn{2}{r|}{Min.} & 0.571 (ja) & 0.026 (iw) & 0.598 (ja) & 0.078 (bn) \\
\bottomrule
\end{tabular}
\caption{\textbf{Semantic clustering performs poorly on an expanded set of supported languages}. The semantic clustering is applied using an expanded set of 17 newly supported languages. LLaMA-3.2 1B generates a text in English, then the translation attack is applied using each of these supported languages as target language. Higher values indicate better robustness. Minimum indicates the worst-case robustness, i.e., the best language for an attack.}
\label{tab:xsir_xkgw_auc_tpr01_translation_llama3_supported}
\end{table}

\begin{table}[ht]
\centering
\small
\setlength{\tabcolsep}{1pt}
\begin{tabular}{cc|cccc}
\toprule
\multicolumn{2}{c}{\textbf{Translation Attack}} & \multicolumn{2}{c}{\textbf{X-SIR} ($\uparrow$)} & \multicolumn{2}{c}{\textbf{X-KGW} ($\uparrow$)} \\
\cmidrule(lr){1-2} \cmidrule(lr){3-4} \cmidrule(lr){5-6}
Type & Language & AUC & TPR@1\% & AUC & TPR@1\% \\
\midrule
\multirow{5}{*}{\makecell[c]{High-\\resource}}
 & fr & 0.804 & 0.249 & 0.852 & 0.466 \\
 & de & 0.833 & 0.399 & 0.850 & 0.484 \\
 & it & 0.829 & 0.336 & 0.870 & 0.478 \\
 & es & 0.811 & 0.319 & 0.869 & 0.506 \\
 & pt & 0.726 & 0.338 & 0.863 & 0.454 \\
\cmidrule[0.3pt](lr){1-6}
\multirow{6}{*}{\makecell[c]{Medium-\\resource}}
 & pl & 0.812 & 0.308 & 0.847 & 0.410 \\
 & nl & 0.847 & 0.355 & 0.882 & 0.592 \\
 & ru & 0.787 & 0.256 & 0.821 & 0.368 \\
 & hi & 0.702 & 0.215 & 0.714 & 0.228 \\
 & ko & 0.792 & 0.276 & 0.822 & 0.422 \\
 & ja & 0.714 & 0.187 & 0.705 & 0.206 \\
\cmidrule[0.3pt](lr){1-6}
\multirow{6}{*}{\makecell[c]{Low-\\resource}}
 & bn & 0.588 & 0.086 & 0.765 & 0.244 \\
 & fa & 0.755 & 0.268 & 0.829 & 0.398 \\
 & vi & 0.772 & 0.238 & 0.802 & 0.328 \\
 & iw & 0.719 & 0.196 & 0.808 & 0.444 \\
 & uk & 0.794 & 0.309 & 0.817 & 0.118 \\
 & ta & 0.561 & 0.067 & 0.789 & 0.316 \\
\cmidrule[0.3pt](lr){1-6}
\multicolumn{2}{r|}{Minimum} & 0.561 (ta) & 0.067 (ta) & 0.705 (ja) & 0.118 (uk) \\
\bottomrule
\end{tabular}
\caption{\textbf{Semantic clustering performs poorly on an expanded set of supported languages}. The semantic clustering is applied using an expanded set of 17 newly supported languages. LLaMAX-3 8B generates a text in English, then the translation attack is applied using each of these supported languages as target language. Higher values indicate better robustness. Minimum indicates the worst-case robustness, i.e., the best language for an attack.}
\label{tab:xsir_xkgw_auc_tpr01_translation_llamax_supported}
\end{table}

\begin{table*}[h!]
\centering
\small
\setlength{\tabcolsep}{3pt}
\begin{tabular}{cc|cccccc}
\toprule
\multicolumn{2}{c}{\textbf{CWRA Attack}} & \multicolumn{2}{c}{\textbf{Aya-23 8B} ($\uparrow$)} & \multicolumn{2}{c}{\textbf{LLaMA-3.2 1B} ($\uparrow$)} & \multicolumn{2}{c}{\textbf{LLaMAX-3 8B} ($\uparrow$)} \\
\cmidrule(lr){1-2} \cmidrule(lr){3-4} \cmidrule(lr){5-6} \cmidrule(lr){7-8}
Type & Language & AUC & TPR@1\% & AUC & TPR@1\% & AUC & TPR@1\% \\
\midrule
\multirow{5}{*}{\makecell[c]{High-\\resource}}
 & fr & 0.831 & 0.201 & 0.898 & 0.421 & 0.845 & 0.345 \\
 & de & 0.820 & 0.198 & 0.902 & 0.394 & 0.841 & 0.361 \\
 & it & 0.817 & 0.196 & 0.872 & 0.331 & 0.825 & 0.315 \\
 & es & 0.819 & 0.200 & 0.859 & 0.281 & 0.816 & 0.320 \\
 & pt & 0.801 & 0.191 & 0.876 & 0.348 & 0.827 & 0.335 \\
\cmidrule[0.3pt](lr){1-8}
\multirow{6}{*}{\makecell[c]{Medium-\\resource}}
 & pl & 0.804 & 0.202 & 0.881 & 0.381 & 0.815 & 0.291 \\
 & nl & 0.859 & 0.265 & 0.899 & 0.434 & 0.834 & 0.367 \\
 & ru & 0.769 & 0.175 & 0.824 & 0.226 & 0.771 & 0.233 \\
 & hi & 0.710 & 0.147 & 0.771 & 0.304 & 0.744 & 0.263 \\
 & ko & 0.769 & 0.178 & 0.854 & 0.340 & 0.805 & 0.299 \\
 & ja & 0.787 & 0.278 & 0.904 & 0.644 & 0.720 & 0.314 \\
\cmidrule[0.3pt](lr){1-8}
\multirow{6}{*}{\makecell[c]{Low-\\resource}}
 & bn & 0.721 & 0.255 & 0.934 & 0.628 & 0.784 & 0.354 \\
 & fa & 0.691 & 0.113 & 0.796 & 0.260 & 0.733 & 0.246 \\
 & vi & 0.781 & 0.180 & 0.865 & 0.325 & 0.805 & 0.301 \\
 & iw & 0.726 & 0.124 & 0.815 & 0.207 & 0.729 & 0.245 \\
 & uk & 0.769 & 0.157 & 0.849 & 0.253 & 0.750 & 0.225 \\
 & ta & 0.819 & 0.405 & 0.917 & 0.516 & 0.740 & 0.391 \\
\cmidrule[0.3pt](lr){1-8}
\multicolumn{2}{r|}{Minimum} & 0.691 (fa) & 0.113 (fa) & 0.771 (hi) & 0.207 (iw) & 0.720 (ja) & 0.225 (uk) \\
\bottomrule
\end{tabular}
\caption{\textbf{Semantic clustering (XSIR) performs inconsistently on an expanded set of supported languages}. The semantic clustering is applied using an expanded set of 17 newly supported languages. A prompt in English is first translated into each target language. Aya-23 8B, LLaMA-3.2 1B, and LLaMAX-3 8B are then prompted with the translated input to generate text in the target language. Finally, the CWRA attack is applied by translating the generated text back into English. Higher values indicate better robustness. Minimum indicates the worst-case robustness, i.e., the best language for an attack.}
\label{tab:xsir_auc_tpr01_cwra_aya_llamax_llama3_supported}
\end{table*}

\subsection{Tokenizer vocabulary analysis}
    \label{app:tokenizer-dicts}

    \begin{figure}[ht]
        \centering
        \includegraphics[width=\linewidth]{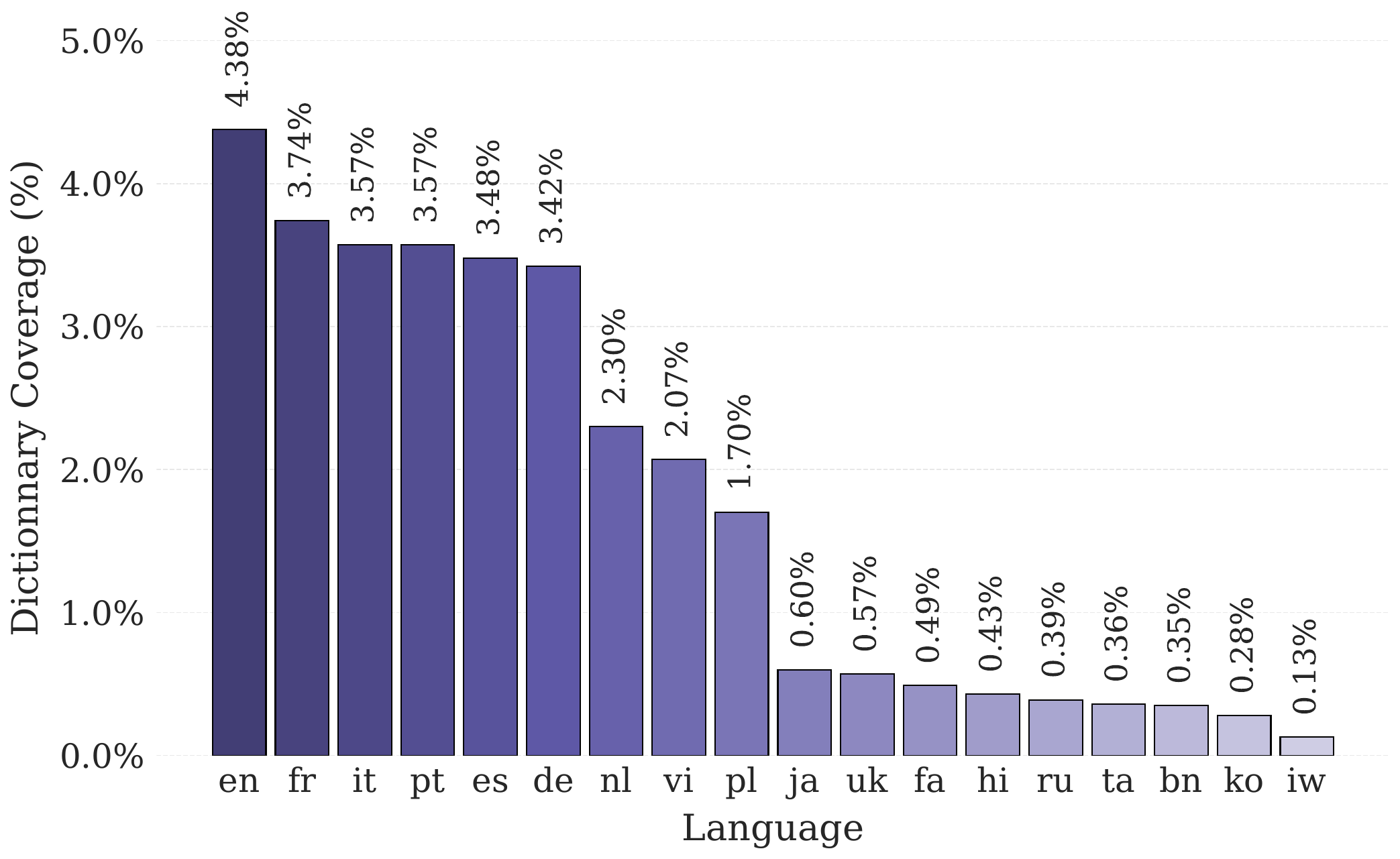}
        \caption{\textbf{Tokenizer vocabulary favours high-resource languages}. Percentage of words in multilingual dictionaries that appear in the tokenizer vocabulary.}
        \label{fig:dictionary_coverage}
    \end{figure}

\subsection{Sub-character Token Distributions}
    \label{app:subchar_dist}

\begin{figure}[htbp]
        \centering
    \begin{subfigure}{0.8\linewidth}
            \centering
            \includegraphics[width=\linewidth]{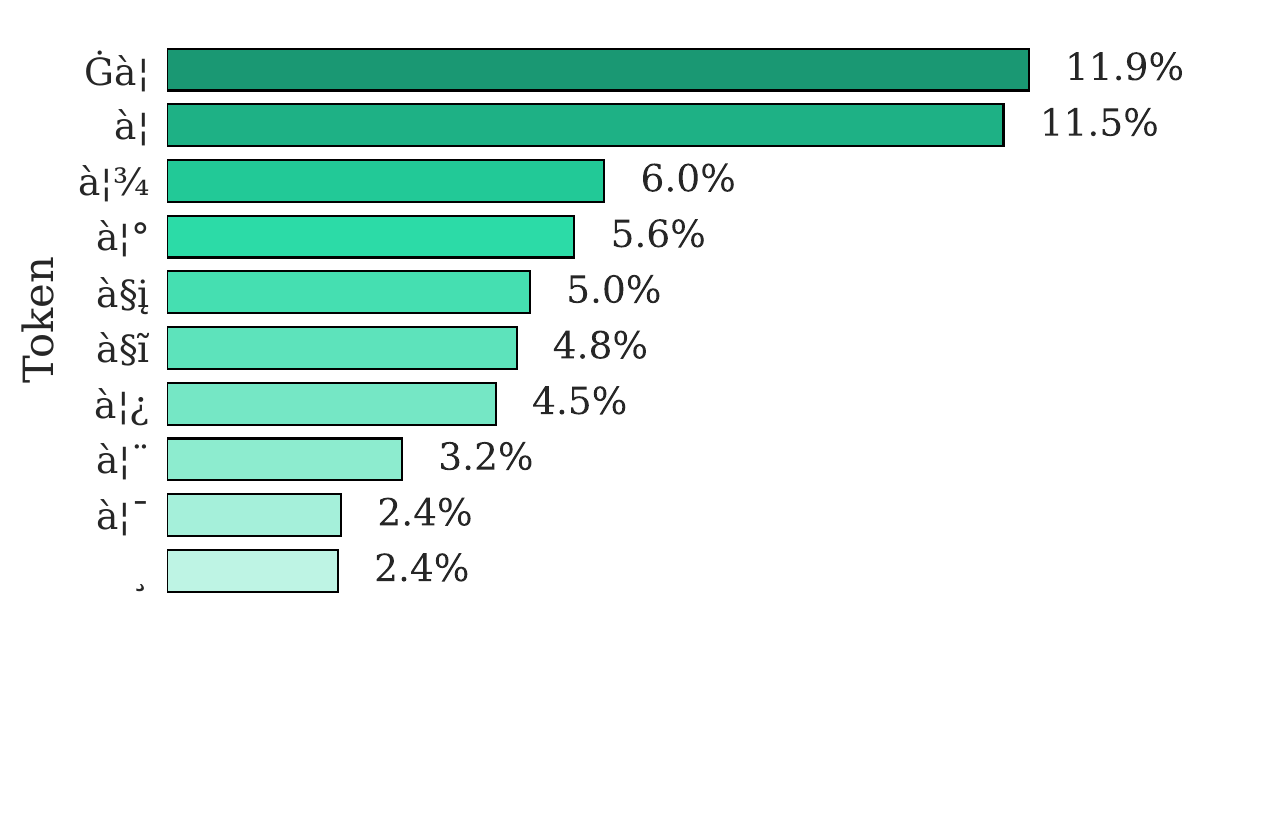}
            \caption{\textbf{bn}}
            \label{fig:bn_aya}
        \end{subfigure}
        \hspace{1em}
        \begin{subfigure}{0.8\linewidth}
            \centering
            \includegraphics[width=\linewidth]{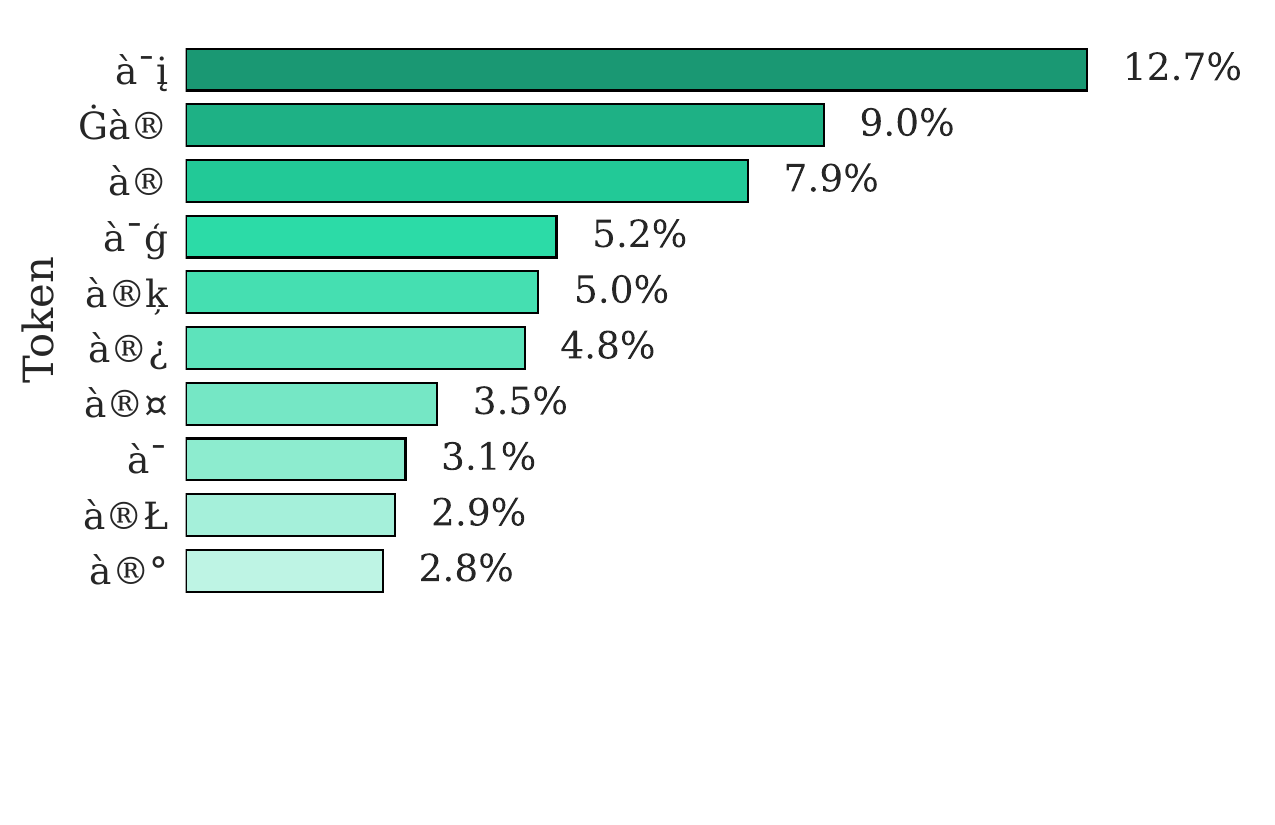}
            \caption{\textbf{ta}}
            \label{fig:ta_aya}
        \end{subfigure}
    
        \caption{\textbf{Tokenization of low-resource languages creates highly concentrated sub-character tokens.} Percentage of top 10 tokens for Bengali (\subref{fig:bn_aya}) and for Tamil (\subref{fig:ta_aya}) using Aya-23 8B.}
        \label{fig:aya_token_distributions}
    \end{figure}

\begin{figure}[htbp]
        \centering
        \begin{subfigure}{0.8\linewidth}
            \centering
            \includegraphics[width=\linewidth]{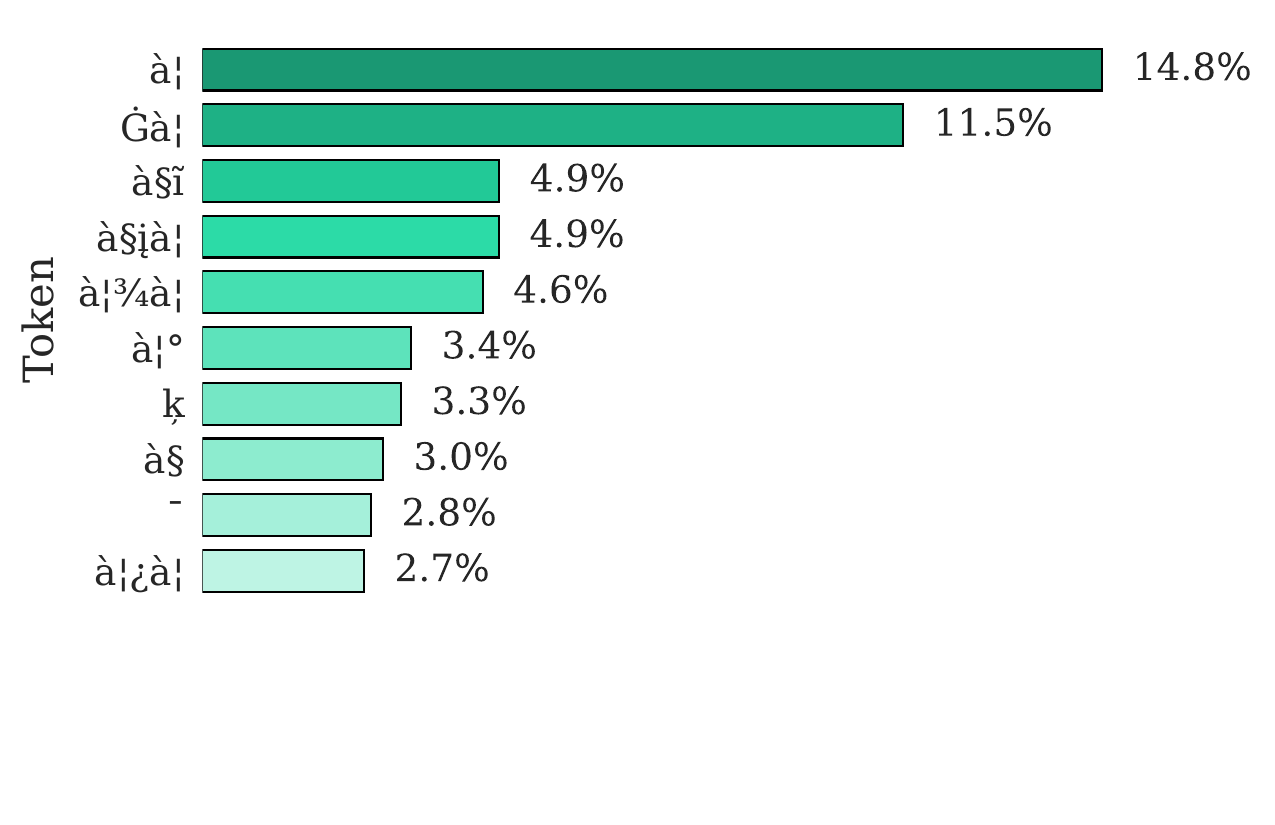}
            \caption{\textbf{bn}}
            \label{fig:bn_llama}
        \end{subfigure}
        \hspace{1em}
        \begin{subfigure}{0.80\linewidth}
            \centering
            \includegraphics[width=\linewidth]{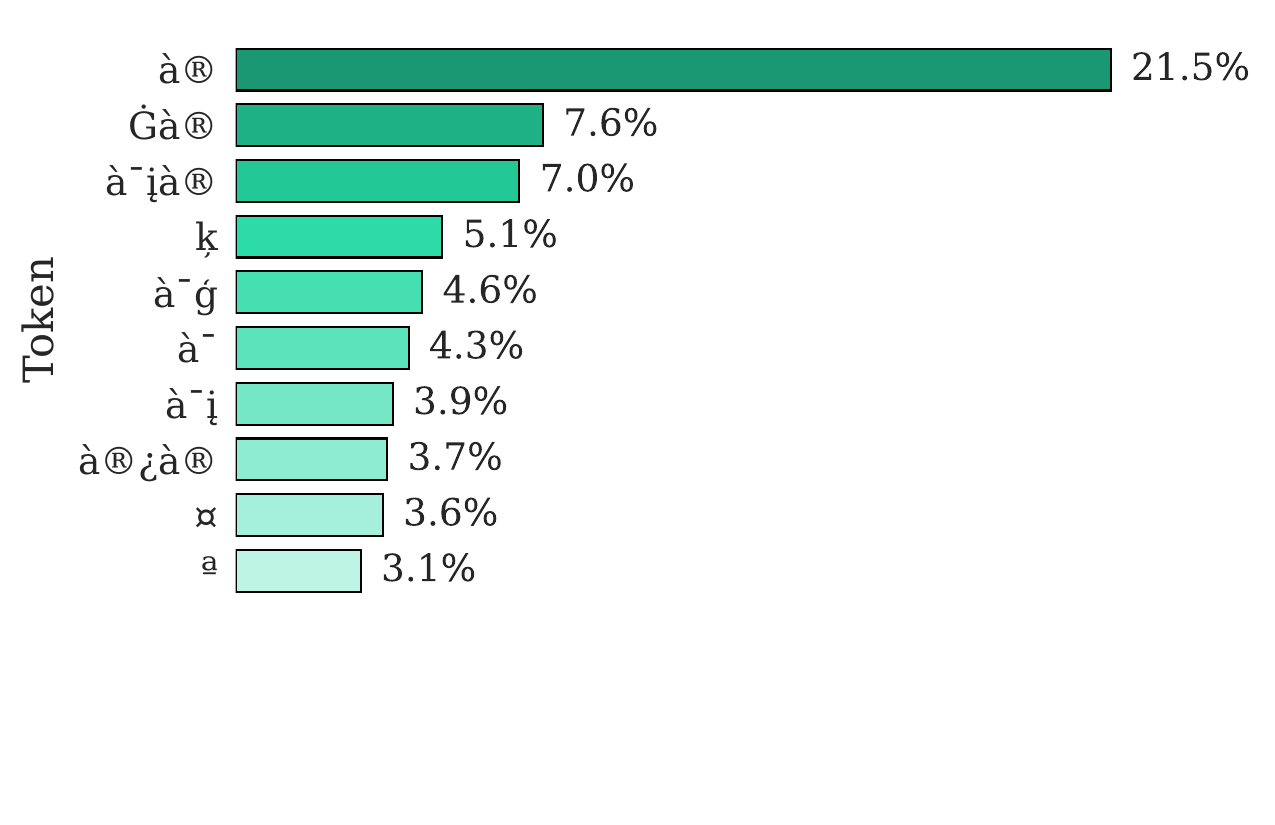}
            \caption{\textbf{ta}}
            \label{fig:ta_llama}
        \end{subfigure}
    
        \caption{\textbf{Tokenization of low-resource languages creates highly concentrated sub-character tokens.} Percentage of top 10 tokens for Bengali (\subref{fig:bn_llama}) and for Tamil (\subref{fig:ta_llama}) using LLaMA-3.2 1B.}
        \label{fig:llama3_token_distributions}
    \end{figure}

    As discussed in \S \ref{subsec:steam_mechanism}, the z-score normalization component is designed to calibrate \textsc{Steam}'s detection mechanism against statistical noise introduced by tokenizer limitations. Figures \ref{fig:aya_token_distributions} and \ref{fig:llama3_token_distributions} show the token distribution for two severely affected low-resource languages, Bengali and Tamil.
    
    \clearpage
    
    \section{Additional Results About STEAM}
    \label{app:steam}

    Tables~\ref{tab:llama3_translation_xsir_steam_grouped} and~\ref{tab:gemma_translation_xsir_steam_grouped} report \textsc{Steam} performance on 17 supported languages for LLaMA-3.2 1B and Gemma4-E4B, complementing the Aya-23 8B results in Table~\ref{tab:aya_translation_xsir_steam_grouped}. Table ~\ref{tab:steam-all-supported} reports \textsc{Steam} performance on all 126 supported languages for Aya-23 8B. Tables~\ref{tab:aya_translation_xsir_steam_unsupported} and~\ref{tab:llama3_translation_xsir_steam_unsupported} report results on 10 unsupported languages for Aya-23 8B and LLaMA-3.2 1B, respectively, complementing the AUC summary in Table~\ref{tab:aya_translation_xsir_steam_unsupported_auc}. Table~\ref{tab:aya_translation_vs_steam_grouped_none_de_ko_bn} reports results under the multistep translation attack. Table~\ref{tab:kgw_llama_norm_summary} shows the impact of per-language z-score normalization on STEAM’s performance

    \begin{table*}[h!]
\centering
\small
\setlength{\tabcolsep}{3pt}
\begin{tabular}{cc|cccc|cccc}
\toprule
 \multicolumn{2}{c}{\textbf{Translation Attack}} 
 & \multicolumn{4}{c}{\textbf{AUC} ($\uparrow$)} 
 & \multicolumn{4}{c}{\textbf{TPR@1\%} ($\uparrow$)} \\
 \cmidrule(lr){1-2} \cmidrule(lr){3-6}  \cmidrule(lr){7-10}
 Type & Language & KGW & X-KGW & X-SIR & STEAM\:\steam & KGW & X-KGW & X-SIR & STEAM\:\steam \\
\midrule
\multirow{5}{*}{\makecell[c]{High-\\resource}}
 & fr & 0.655 & 0.719 & 0.702 & \textbf{0.969} & 0.052 & 0.166 & 0.085 & \textbf{0.618} \\
 & de & 0.649 & 0.752 & 0.708 & \textbf{0.975} & 0.090 & 0.186 & \red{0.067} & \textbf{0.530} \\
 & it & 0.617 & 0.750 & 0.712 & \textbf{0.977} & 0.118 & 0.230 & \red{0.111} & \textbf{0.620} \\
 & es & 0.616 & 0.724 & 0.703 & \textbf{0.969} & 0.122 & 0.222 & \red{0.089} & \textbf{0.522} \\
 & pt & 0.651 & 0.747 & 0.726 & \textbf{0.970} & 0.096 & 0.206 & 0.102 & \textbf{0.476} \\
\cmidrule[0.3pt](lr){1-10}
\multirow{6}{*}{\makecell[c]{Medium-\\resource}}
 & pl & 0.622 & 0.703 & 0.657 & \textbf{0.975} & 0.088 & 0.188 & \red{0.065} & \textbf{0.526} \\
 & nl & 0.719 & 0.787 & 0.722 & \textbf{0.973} & 0.126 & 0.252 & \red{0.091} & \textbf{0.580} \\
 & ru & 0.629 & 0.656 & 0.635 & \textbf{0.970} & 0.052 & 0.100 & 0.075 & \textbf{0.482} \\
 & hi & 0.568 & 0.620 & 0.611 & \textbf{0.974} & 0.048 & 0.084 & \red{0.037} & \textbf{0.598} \\
 & ko & 0.625 & 0.701 & 0.673 & \textbf{0.971} & 0.068 & 0.154 & \red{0.055} & \textbf{0.416 }\\
 & ja & 0.578 & 0.598 & \red{0.571} & \textbf{0.973} & 0.048 & 0.110 & \red{0.042} & \textbf{0.482} \\
\cmidrule[0.3pt](lr){1-10}
\multirow{6}{*}{\makecell[c]{Low-\\resource}}
 & bn & 0.574 & 0.701 & 0.825 & \textbf{0.967} & 0.020 & 0.078 & \textbf{0.509} & 0.496 \\
 & fa & 0.586 & 0.673 & \red{0.584} & \textbf{0.979} & 0.082 & 0.086 & \red{0.055} & \textbf{0.664} \\
 & vi & 0.658 & 0.722 & 0.691 & \textbf{0.975} & 0.082 & 0.186 & 0.084 & \textbf{0.458} \\
 & iw & 0.495 & 0.702 & 0.622 & \textbf{0.966} & 0.042 & 0.134 & \red{0.026} & \textbf{0.502 }\\
 & uk & 0.629 & 0.725 & \red{0.613} & \textbf{0.973} & 0.084 & 0.118 & \red{0.064} & \textbf{0.482} \\
 & ta & 0.877 & \red{0.672} & \red{0.749} & \textbf{0.977} & 0.272 & \red{0.108} & \red{0.095} & \textbf{0.460} \\
\bottomrule
\end{tabular}
\caption{\textbf{\textsc{Steam}\:\steam consistently and substantially outperforms semantic clustering.} Watermark strength (AUC and TPR@1\%) of multilingual watermarking techniques with 17 supported languages and LLaMA-3.2 1B. Red indicates that the defence reduces robustness (lower than the undefended KGW baseline). Bolded is best.}
\label{tab:llama3_translation_xsir_steam_grouped}
\end{table*}

\begin{table*}[h!]
\centering
\small
\setlength{\tabcolsep}{2pt}
\begin{tabular}{ccccccccc}
\toprule 
\multirow{2}{*}{\makecell[b]{\textbf{New}\\\textbf{Lang.}}} & \multicolumn{4}{c}{\textbf{AUC} ($\uparrow$)} 
 & \multicolumn{4}{c}{\textbf{TPR@1\%} ($\uparrow$)} \\
 \cmidrule(lr){2-5} \cmidrule(lr){6-9}
 & KGW & X-KGW & X-SIR & STEAM\:\steam & KGW & X-KGW & X-SIR & STEAM\:\steam \\
\midrule
it & 0.733 & 0.772 & 0.796 & \textbf{0.966} & 0.202 & 0.238 & \red{0.177} & \textbf{0.494} \\
es & 0.717 & 0.807 & 0.754 & \textbf{0.967} & 0.232 & \red{0.230} & \red{0.155} & \textbf{0.548} \\
pt & 0.732 & 0.792 & 0.775 & \textbf{0.971} & 0.242 & 0.286 & \red{0.133} & \textbf{0.466} \\
pl & 0.730 & 0.762 & 0.749 & \textbf{0.960} & 0.248 & \red{0.236} & \red{0.127} & \textbf{0.344} \\
nl & 0.768 & 0.808 & 0.776 & \textbf{0.966} & 0.286 & 0.314 & \red{0.164} & \textbf{0.358} \\
hr & 0.706 & 0.757 & 0.726 & \textbf{0.965} & 0.194 & 0.210 & \red{0.124} & \textbf{0.362} \\
cs & 0.717 & 0.754 & 0.773 & \textbf{0.974} & 0.212 & 0.254 & \red{0.111} & \textbf{0.554} \\
da & 0.713 & 0.764 & 0.734 & \textbf{0.971} & 0.196 & 0.266 & \red{0.161} & \textbf{0.448} \\
ko & 0.732 & 0.754 & \red{0.729} & \textbf{0.961} & 0.220 & 0.226 & \red{0.136} & \textbf{0.322} \\
ar & 0.689 & 0.765 & \red{0.687} & \textbf{0.971} & 0.186 & \red{0.168} & \red{0.093} & \textbf{0.588} \\
\bottomrule
\end{tabular}
\caption{\textbf{\textsc{Steam}\:\steam outperforms semantic clustering methods on unsupported languages.} Watermark strength (AUC and TPR@1\%) of multilingual watermarking techniques with 10 unsupported languages and Aya-23 8B. Red indicates that the defence reduces robustness (lower than the undefended KGW baseline). Bolded is best.}
\label{tab:aya_translation_xsir_steam_unsupported}
\end{table*}

\begin{table*}[h!]
\centering
\small
\setlength{\tabcolsep}{2pt}
\begin{tabular}{ccccccccc}
\toprule 
\multirow{2}{*}{\makecell[b]{\textbf{New}\\\textbf{Lang.}}} & \multicolumn{4}{c}{\textbf{AUC} ($\uparrow$)} 
 & \multicolumn{4}{c}{\textbf{TPR@1\%} ($\uparrow$)} \\
 \cmidrule(lr){2-5} \cmidrule(lr){6-9}
 & KGW & X-KGW & X-SIR & STEAM\:\steam & KGW & X-KGW & X-SIR & STEAM\:\steam \\
\midrule
it & 0.620 & 0.760 & 0.699 & \textbf{0.964} & 0.108 & 0.212 & \red{0.069} & \textbf{0.350} \\
es & 0.616 & 0.744 & 0.665 & \textbf{0.961} & 0.122 & 0.222 & \red{0.076} & \textbf{0.358} \\
pt & 0.652 & 0.722 & \red{0.641} & \textbf{0.966} & 0.096 & 0.152 & \red{0.059} & \textbf{0.408} \\
pl & 0.617 & 0.677 & 0.679 & \textbf{0.965} & 0.088 & 0.144 & \red{0.069} & \textbf{0.310} \\
nl & 0.714 & 0.781 & 0.754 & \textbf{0.957} & 0.112 & 0.244 & \red{0.095} & \textbf{0.258} \\
hr & 0.611 & 0.733 & 0.660 & \textbf{0.966} & 0.078 & 0.162 & \red{0.066} & \textbf{0.448} \\
cs & 0.655 & 0.759 & \red{0.650} & \textbf{0.956} & 0.072 & 0.190 & \red{0.064} & \textbf{0.228} \\
da & 0.655 & 0.765 & 0.675 & \textbf{0.962} & 0.080 & 0.196 & 0.093 & \textbf{0.346} \\
ko & 0.623 & 0.672 & 0.673 & \textbf{0.961} & 0.066 & 0.124 & \red{0.062} & \textbf{0.300} \\
ar & 0.635 & 0.704 & 0.655 & \textbf{0.963} & 0.110 & 0.168 & \red{0.055} & \textbf{0.286} \\
\bottomrule
\end{tabular}
\caption{\textbf{\textsc{Steam}\:\steam outperforms semantic clustering methods on unsupported languages in almost all cases.} Watermark strength (AUC and TPR@1\%) of multilingual watermarking techniques with 10 unsupported languages and LLaMA-3.2 1B. Bold marks the best per row; red indicates a defended score lower than the KGW baseline.}
\label{tab:llama3_translation_xsir_steam_unsupported}
\end{table*}
    \begin{table*}[h!]
\centering
\small
\setlength{\tabcolsep}{4pt}
\begin{tabular}{l cccc cccc}
\toprule
& \multicolumn{4}{c}{\textbf{AUC} ($\uparrow$)} & \multicolumn{4}{c}{\textbf{TPR@1\%} ($\uparrow$)} \\
\cmidrule(lr){2-5} \cmidrule(lr){6-9}
\multirow{1}{*}{Lang.} & KGW & X-KGW & X-SIR & STEAM \steam & KGW & X-KGW & X-SIR & STEAM \steam \\
\midrule
\texttt{fr} & 0.722 & 0.744 & \red{0.450} & \textbf{0.978} & 0.250 & \red{0.222} & \red{0.000} & \textbf{0.604} \\
\texttt{de} & 0.740 & 0.787 & \red{0.493} & \textbf{0.984} & 0.284 & 0.306 & \red{0.000} & \textbf{0.768} \\
\texttt{it} & 0.745 & 0.810 & \red{0.552} & \textbf{0.984} & 0.270 & \red{0.250} & \red{0.000} & \textbf{0.654} \\
\texttt{es} & 0.745 & 0.827 & \red{0.521} & \textbf{0.985} & 0.356 & \red{0.336} & \red{0.000} & \textbf{0.668} \\
\texttt{pt} & 0.752 & 0.834 & \red{0.556} & \textbf{0.982} & 0.252 & \red{0.236} & \red{0.000} & \textbf{0.700} \\
\texttt{pl} & 0.743 & 0.767 & \red{0.535} & \textbf{0.983} & 0.276 & \red{0.204} & \red{0.000} & \textbf{0.666} \\
\texttt{nl} & 0.771 & 0.798 & \red{0.545} & \textbf{0.982} & 0.250 & 0.300 & \red{0.000} & \textbf{0.698} \\
\texttt{ru} & 0.730 & \red{0.724} & \red{0.546} & \textbf{0.975} & 0.226 & \red{0.140} & \red{0.000} & \textbf{0.520} \\
\texttt{hi} & 0.638 & 0.728 & \red{0.483} & \textbf{0.980} & 0.174 & 0.196 & \red{0.000} & \textbf{0.588} \\
\texttt{ko} & 0.691 & 0.802 & \red{0.558} & \textbf{0.981} & 0.176 & \red{0.168} & \red{0.000} & \textbf{0.646} \\
\texttt{ja} & 0.699 & 0.757 & \red{0.533} & \textbf{0.973} & 0.204 & \red{0.182} & \red{0.000} & \textbf{0.554} \\
\texttt{bn} & 0.682 & 0.691 & \red{0.442} & \textbf{0.977} & 0.216 & \red{0.176} & \red{0.000} & \textbf{0.610} \\
\texttt{fa} & 0.719 & 0.793 & \red{0.583} & \textbf{0.986} & 0.266 & \red{0.250} & \red{0.000} & \textbf{0.718} \\
\texttt{vi} & 0.756 & \red{0.754} & \red{0.482} & \textbf{0.981} & 0.304 & \red{0.260} & \red{0.000} & \textbf{0.652} \\
\texttt{iw} & 0.705 & \red{0.669} & \red{0.449} & \textbf{0.975} & 0.236 & \red{0.146} & \red{0.000} & \textbf{0.592} \\
\texttt{uk} & 0.726 & 0.739 & \red{0.509} & \textbf{0.980} & 0.262 & \red{0.182} & \red{0.000} & \textbf{0.666} \\
\texttt{ta} & 0.666 & 0.715 & \red{0.469} & \textbf{0.981} & 0.192 & \red{0.162} & \red{0.000} & \textbf{0.672} \\
\bottomrule
\end{tabular}
\caption{\textbf{\textsc{Steam} \steam outperforms all baselines across all 17 languages on Gemma4-E4B.} AUC and TPR@1\% for KGW, X-KGW, X-SIR, and \textsc{Steam} \steam. X-SIR collapses to TPR@1\%~$=$~0.000 on every language. Red indicates values fall below the KGW baseline. Bolded is best.}
\label{tab:gemma_translation_xsir_steam_grouped}
\end{table*}
    \begin{table*}[ht]
\centering
\small
\setlength{\tabcolsep}{5pt}
\begin{tabular}{l cc | l cc | l cc}
\toprule

\textbf{Lang.} & \textbf{AUC} ($\uparrow$) & \textbf{TPR@1\%} ($\uparrow$) &
\textbf{Lang.} & \textbf{AUC} ($\uparrow$) & \textbf{TPR@1\% }($\uparrow$) &
\textbf{Lang.} & \textbf{AUC} ($\uparrow$) & \textbf{TPR@1\%}($\uparrow$) \\

\midrule

af & 0.988 & 0.756 & ig & 0.974 & 0.592 & ru & 0.971 & 0.576 \\
sq & 0.981 & 0.718 & ilo & 0.982 & 0.678 & sm & 0.983 & 0.668 \\
am & 0.976 & 0.588 & id & 0.979 & 0.644 & sa & 0.981 & 0.684 \\
ar & 0.984 & 0.654 & ga & 0.980 & 0.618 & gd & 0.989 & 0.802 \\
hy & 0.980 & 0.700 & it & 0.978 & 0.530 & sr & 0.981 & 0.618 \\
as & 0.969 & 0.568 & ja & 0.975 & 0.498 & st & 0.979 & 0.618 \\
ay & 0.985 & 0.672 & jw & 0.980 & 0.610 & sn & 0.981 & 0.562 \\
az & 0.977 & 0.626 & kn & 0.977 & 0.656 & sd & 0.984 & 0.718 \\
bm & 0.982 & 0.674 & kk & 0.973 & 0.502 & si & 0.975 & 0.508 \\
eu & 0.980 & 0.608 & km & 0.979 & 0.582 & sk & 0.979 & 0.630 \\
be & 0.980 & 0.694 & rw & 0.983 & 0.632 & sl & 0.979 & 0.572 \\
bn & 0.978 & 0.604 & ko & 0.968 & 0.460 & so & 0.985 & 0.694 \\
bs & 0.980 & 0.584 & kri & 0.984 & 0.712 & es & 0.976 & 0.580 \\
bg & 0.982 & 0.610 & ku & 0.977 & 0.638 & su & 0.982 & 0.606 \\
ca & 0.982 & 0.706 & ckb & 0.977 & 0.608 & sw & 0.984 & 0.680 \\
ny & 0.983 & 0.670 & ky & 0.975 & 0.576 & sv & 0.981 & 0.686 \\
zh-CN & 0.976 & 0.544 & lo & 0.984 & 0.670 & tg & 0.977 & 0.610 \\
co & 0.974 & 0.590 & la & 0.975 & 0.618 & ta & 0.967 & 0.504 \\
hr & 0.978 & 0.670 & lv & 0.981 & 0.624 & tt & 0.978 & 0.586 \\
cs & 0.984 & 0.700 & ln & 0.976 & 0.562 & te & 0.976 & 0.612 \\
da & 0.983 & 0.718 & lt & 0.982 & 0.722 & th & 0.977 & 0.616 \\
dv & 0.976 & 0.608 & lg & 0.977 & 0.598 & ti & 0.982 & 0.660 \\
nl & 0.982 & 0.612 & lb & 0.988 & 0.784 & ts & 0.977 & 0.610 \\
eo & 0.979 & 0.616 & mk & 0.976 & 0.564 & tr & 0.983 & 0.642 \\
et & 0.987 & 0.746 & mai & 0.977 & 0.610 & tk & 0.975 & 0.538 \\
ee & 0.979 & 0.588 & mg & 0.976 & 0.560 & ak & 0.978 & 0.616 \\
tl & 0.983 & 0.700 & ms & 0.982 & 0.610 & uk & 0.979 & 0.542 \\
fi & 0.976 & 0.644 & ml & 0.979 & 0.638 & ur & 0.986 & 0.736 \\
fr & 0.976 & 0.582 & mt & 0.978 & 0.590 & ug & 0.802 & 0.214 \\
fy & 0.983 & 0.656 & mi & 0.979 & 0.564 & uz & 0.977 & 0.616 \\
gl & 0.972 & 0.558 & mr & 0.979 & 0.584 & vi & 0.976 & 0.626 \\
ka & 0.989 & 0.756 & mni-Mtei & 0.979 & 0.650 & cy & 0.984 & 0.684 \\
de & 0.973 & 0.622 & lus & 0.976 & 0.608 & xh & 0.982 & 0.596 \\
el & 0.981 & 0.700 & mn & 0.976 & 0.584 & yo & 0.980 & 0.680 \\
gu & 0.829 & 0.284 & my & 0.977 & 0.582 & zu & 0.988 & 0.724 \\
ht & 0.981 & 0.692 & ne & 0.970 & 0.570 & pa & 0.968 & 0.456 \\
ha & 0.979 & 0.702 & no & 0.987 & 0.742 & bho & 0.973 & 0.562 \\
haw & 0.980 & 0.616 & or & 0.980 & 0.610 & ceb & 0.976 & 0.578 \\
iw & 0.974 & 0.578 & fa & 0.979 & 0.664 & nso & 0.984 & 0.682 \\
hi & 0.978 & 0.644 & pl & 0.975 & 0.526 & yi & 0.983 & 0.608 \\
hu & 0.979 & 0.660 & pt & 0.977 & 0.538 &  &  &  \\
is & 0.976 & 0.592 & ro & 0.981 & 0.716 &  &  &  \\

\bottomrule
\end{tabular}
\caption{\textbf{\textsc{Steam} \steam maintains robustness under all supported languages.} Watermark strength (AUC and TPR@1\%) and Aya-23 8B.}
\label{tab:steam-all-supported}
\end{table*}


    \begin{table}[ht]
\centering
\small
\setlength{\tabcolsep}{3pt}
\begin{tabular}{cc|cc}
\toprule
\multicolumn{2}{c}{\textbf{Two-Step Translation Attack}} & \multicolumn{2}{c}{\textbf{STEAM\:\steam}} \\
\cmidrule(lr){1-2} \cmidrule(lr){3-4}
Language 1 & Language 2 & AUC\:$\uparrow$ & TPR@1\%\:$\uparrow$ \\
\midrule
\multirow{4}{*}{\makecell[c]{High-\\resource}}
 & None & \textbf{0.976} & \textbf{0.570} \\
 & de   & 0.899 & 0.343 \\
 & ko   & 0.833 & 0.227 \\
 & bn   & 0.864 & 0.252 \\
\cmidrule[0.3pt](lr){1-4}
\multirow{4}{*}{\makecell[c]{Medium-\\resource}}
 & None & \textbf{0.975} & \textbf{0.553} \\
 & de   & 0.910 & 0.349 \\
 & ko   & 0.890 & 0.239 \\
 & bn   & 0.872 & 0.396 \\
\cmidrule[0.3pt](lr){1-4}
\multirow{4}{*}{\makecell[c]{Low-\\resource}}
 & None & \textbf{0.976} & \textbf{0.586} \\
 & de   & 0.875 & 0.164 \\
 & ko   & 0.938 & 0.252 \\
 & bn   & 0.871 & 0.325 \\
\bottomrule
\end{tabular}
\caption{\textbf{\textsc{Steam}\: \steam remains robust under multi-step attacks.} Aya-23 8B generates text in English that is translated to the 17 supported languages (\textit{Language 1}). A second translation step is then applied using \textit{Language~2} to compute the AUC and TPR@1\%. None indicates the single-step translation baseline.}
\label{tab:aya_translation_vs_steam_grouped_none_de_ko_bn}
\end{table}

        \subsection{Inference Cost}
            \label{app:inference-cost}
\textsc{Steam} introduces no additional cost at generation time, as text is produced by the unmodified watermarking scheme (e.g., vanilla KGW). The only cost is at detection time, where \textsc{Steam} requires translating the suspect text into back-translation languages (Table~\ref{tab:translation_cost}). Google Translate is free, while high-quality LLM-based translators such as DeepSeek-V3.2-Exp cost \$0.003 per text, making \textsc{Steam} practical even at scale.
    
    \begin{table}[H]
\centering
\small
\setlength{\tabcolsep}{3pt}
\begin{tabular}{l r}
\toprule
\textbf{Model} & \textbf{Cost per text (\$)} \\
\midrule
Google Translate & 0.000 \\
DeepSeek-V3.2-Exp & 0.060 \\
GPT-5 mini & 0.180 \\
Gemini 2.5 Flash & 0.220 \\
\bottomrule
\end{tabular}
\caption{\textbf{\textsc{Steam} \steam requires minimal cost at detection time}. Total cost in US dollars for 20 back-translations per text (the maximum number of evaluated languages by BO). Google Translate is free. \textsc{Steam} adds no cost at generation time.}
\label{tab:translation_cost}
\end{table}

  \subsection{Detection-Time Cost}
  \label{app:detection-cost}
We measure wall-clock latency on 20 watermarked texts using Google Translate at BO budget 20. The mean per-text latency is 17.85\,s, decomposed as follows: defender back-translation (4.14\,s), local KGW detection (2.76\,s), GP refitting (10.64\,s), and acquisition scoring (0.16\,s). GP refitting dominates but runs entirely on CPU, requires no GPU, and parallelises trivially across suspect texts, allowing deployers to saturate available cores at negligible marginal cost. Detection generally does not need to operate in real time, so the latency should be a minor practical issue.

    \subsection{Non-English Source Languages}
    \label{app:source-language}

The main results watermark English text and attack it by translating out of
English. We invert that setting. The watermarked text is generated directly in a
non-English source language, and each text is then attacked by a translation into
a language drawn uniformly at random from a 40-language pool spanning all three
resource tiers. The detector is therefore never tuned to a single known attack
language. Across the 17 source languages in
Table~\ref{tab:aya_source_lang_steam_grouped}, \textsc{Steam} \steam holds a mean
TPR@1\% of 0.504, ranging from 0.321 for Ukrainian to 0.638 for Japanese. Mean AUC is 0.971 over the languages measured, and the weakest language (Ukrainian) scored 0.944.

    \begin{table}[H]
\centering
\small
\setlength{\tabcolsep}{3pt}

\begin{tabular}{cc|cc}
\toprule
 \multicolumn{2}{c}{\textbf{Source Language}}
 & \multicolumn{2}{c}{\textbf{STEAM\:\steam}} \\
 \cmidrule(lr){1-2} \cmidrule(lr){3-4}
 Type & Language & AUC ($\uparrow$) & TPR@1\% ($\uparrow$) \\
\midrule
\multirow{5}{*}{\makecell[c]{High-\\resource}}
 & fr & 0.970 & 0.436 \\
 & de & 0.966 & 0.392 \\
 & it & 0.971 & 0.458 \\
 & es & 0.973 & 0.610 \\
 & pt & 0.972 & 0.488 \\
\cmidrule[0.3pt](lr){1-4}
\multirow{6}{*}{\makecell[c]{Medium-\\resource}}
 & pl & 0.976 & 0.446 \\
 & nl & 0.982 & 0.628 \\
 & ru & 0.971 & 0.518 \\
 & hi & 0.969 & 0.544 \\
 & ko & 0.975 & 0.552 \\
 & ja & 0.981 & 0.638 \\
\cmidrule[0.3pt](lr){1-4}
\multirow{6}{*}{\makecell[c]{Low-\\resource}}
 & bn & 0.949 & 0.460 \\
 & fa & 0.970 & 0.493 \\
 & vi & 0.975 & 0.579 \\
 & iw & 0.970 & 0.587 \\
 & uk & 0.944 & 0.321 \\
 & ta & 0.966 & 0.413 \\
\bottomrule
\end{tabular}
\caption{\textbf{STEAM \steam recovers watermark strength when the watermarked text is generated in a non-English source language.} Watermark strength (AUC and TPR@1\%) with Aya-23 8B, where each text is attacked by a translation into a language drawn at random from a 40-language pool.}
\label{tab:aya_source_lang_steam_grouped}
\vspace{-1em}
\end{table}

    \subsection{Text Length Analysis}
    \label{app:text-length-analysis}

We analyse the impact of text length on watermark detection. For each language, we split texts into three equal-sized bins (short, medium, and long) based on percentiles of token length (Table~\ref{tab:text_length_aya} and Table~\ref{tab:text_length_llama}). \textsc{Steam} maintains strong detection across all length categories, with average AUC above 0.97 across all text lengths for Aya-23 and LLaMA-3.2 models. Even in the most challenging case (Hebrew, long texts, LLaMA-3.2 1B), \textsc{Steam} achieves an AUC of 0.899.

\subsection{Back-Translation with a Local Open-Weights Translator}
\label{app:local_translator}

\textsc{Steam} \steam does not depend on an external translation API, since a
small open-weights translator run locally is sufficient for reliable detection.
We repeat the main evaluation of \S\ref{subsec:steam_evaluation} with NLLB-200 distilled 600M \citep{nllbteam2022} as the defender's translation model and
keep every other component fixed. Aya-23 8B generates English watermarked text
and the translation attack targets each of the 17 supported languages. Detection
holds throughout (Table~\ref{tab:aya_source_lang_nllb_grouped}). \textsc{Steam}
reaches a mean AUC of 0.907 in a narrow band across the resource spectrum, from
0.897 for Japanese to 0.913 for Portuguese, and a mean TPR@1\% of 0.233, again
weakest for Japanese (0.194) and strongest for Portuguese (0.259). Both metrics
remain well above the undefended KGW baseline of 0.698 AUC and 0.186 TPR@1\%
(Table~\ref{tab:aya_translation_xsir_steam_grouped}). The cost of replacing the commercial translator
falls mainly in the low false-positive regime. Relative to Google Translate,
mean AUC declines from 0.975 to 0.907 while mean TPR@1\% declines from 0.570 to
0.233. A defender constrained by budget or privacy can therefore run
\textsc{Steam} entirely locally, at the price of a lower detection rate at a
fixed 1\% FPR.

    \begin{table}[H]
\centering
\small
\setlength{\tabcolsep}{3pt}
\begin{tabular}{cc|cc}
\toprule
 \multicolumn{2}{c}{\textbf{Attack Language}}
 & \multicolumn{2}{c}{\textbf{NLLB}} \\
 \cmidrule(lr){1-2} \cmidrule(lr){3-4}
 Type & Language & AUC ($\uparrow$) & TPR@1\% ($\uparrow$) \\
\midrule
\multirow{5}{*}{\makecell[c]{High-\\resource}}
 & fr & 0.912 & 0.248 \\
 & de & 0.910 & 0.250 \\
 & it & 0.911 & 0.236 \\
 & es & 0.909 & 0.228 \\
 & pt & 0.913 & 0.259 \\
\cmidrule[0.3pt](lr){1-4}
\multirow{6}{*}{\makecell[c]{Medium-\\resource}}
 & pl & 0.908 & 0.221 \\
 & nl & 0.910 & 0.231 \\
 & ru & 0.909 & 0.233 \\
 & hi & 0.911 & 0.247 \\
 & ko & 0.899 & 0.195 \\
 & ja & 0.897 & 0.194 \\
\cmidrule[0.3pt](lr){1-4}
\multirow{6}{*}{\makecell[c]{Low-\\resource}}
 & bn & 0.900 & 0.232 \\
 & fa & 0.906 & 0.234 \\
 & vi & 0.907 & 0.237 \\
 & iw & 0.903 & 0.245 \\
 & uk & 0.904 & 0.244 \\
 & ta & 0.909 & 0.222 \\
\bottomrule
\end{tabular}
\caption{\textbf{\textsc{Steam} \steam remains effective when the defender back-translates with a small local open-weights model.} Watermark strength (AUC and TPR@1\%) with Aya-23 8B and NLLB-200 as the defender's translation model.}
\label{tab:aya_source_lang_nllb_grouped}
\vspace{-1em}
\end{table}

    \begin{table*}[h!]
\centering
\small
\setlength{\tabcolsep}{3pt}
\begin{tabular}{cc | ccc | ccc}
\toprule
\multicolumn{2}{c}{\textbf{Translation Attack}} & \multicolumn{3}{c}{\textbf{AUC ($\uparrow$)}} & \multicolumn{3}{c}{\textbf{TPR@1\% ($\uparrow$)}} \\
\cmidrule(lr){1-2} \cmidrule(lr){3-5} \cmidrule(lr){6-8}
Type & Language & Short & Medium & Long & Short & Medium & Long \\
\midrule
\multirow{5}{*}{\shortstack{High-\\resource}}
& fr & 0.994 & 0.971 & 0.951 & 0.882 & 0.561 & 0.319 \\
& de & 0.988 & 0.979 & 0.948 & 0.827 & 0.619 & 0.451 \\
& it & 0.994 & 0.984 & 0.953 & 0.835 & 0.582 & 0.317 \\
& es & 0.992 & 0.970 & 0.958 & 0.856 & 0.444 & 0.378 \\
& pt & 0.993 & 0.975 & 0.956 & 0.810 & 0.464 & 0.323 \\
\midrule
\multirow{6}{*}{\shortstack{Medium-\\resource}}
& pl & 0.987 & 0.964 & 0.975 & 0.754 & 0.408 & 0.433 \\
& nl & 0.993 & 0.985 & 0.963 & 0.835 & 0.641 & 0.325 \\
& ru & 0.987 & 0.973 & 0.949 & 0.753 & 0.500 & 0.317 \\
& hi & 0.996 & 0.976 & 0.960 & 0.893 & 0.562 & 0.344 \\
& ko & 0.988 & 0.982 & 0.923 & 0.687 & 0.612 & 0.220 \\
& ja & 0.988 & 0.984 & 0.951 & 0.717 & 0.653 & 0.242 \\
\midrule
\multirow{6}{*}{\shortstack{Low-\\resource}}
& bn & 0.989 & 0.979 & 0.963 & 0.789 & 0.575 & 0.250 \\
& fa & 0.993 & 0.980 & 0.958 & 0.856 & 0.549 & 0.494 \\
& vi & 0.987 & 0.983 & 0.959 & 0.789 & 0.643 & 0.282 \\
& iw & 0.990 & 0.970 & 0.956 & 0.826 & 0.503 & 0.372 \\
& uk & 0.992 & 0.976 & 0.970 & 0.713 & 0.465 & 0.368 \\
& ta & 0.992 & 0.952 & 0.948 & 0.840 & 0.389 & 0.195 \\
\bottomrule
\end{tabular}
\caption{\textbf{\textsc{Steam} \steam maintains strong AUC across all three text lengths}. Watermark strength (AUC and TPR@1\%) for short, medium, and long texts (bottom, middle, and top third by token count) using Aya-23 8B.}
\label{tab:text_length_aya}
\end{table*}

\begin{table*}[h!]
\centering
\small
\setlength{\tabcolsep}{3pt}
\begin{tabular}{cc | ccc | ccc}
\toprule
\multicolumn{2}{c}{\textbf{Translation Attack}} & \multicolumn{3}{c}{\textbf{AUC} ($\uparrow$)} & \multicolumn{3}{c}{\textbf{TPR@1\% ($\uparrow$)}} \\
\cmidrule(lr){1-2} \cmidrule(lr){3-5} \cmidrule(lr){6-8}
Type & Language & Short & Medium & Long & Short & Medium & Long \\
\midrule
\multirow{5}{*}{\shortstack{High-\\resource}}
& fr & 0.994 & 0.960 & 0.951 & 0.922 & 0.243 & 0.390 \\
& de & 0.996 & 0.963 & 0.957 & 0.861 & 0.471 & 0.146 \\
& it & 0.996 & 0.980 & 0.947 & 0.899 & 0.583 & 0.018 \\
& es & 0.991 & 0.981 & 0.930 & 0.826 & 0.653 & 0.110 \\
& pt & 0.996 & 0.957 & 0.945 & 0.922 & 0.254 & 0.146 \\
\midrule
\multirow{6}{*}{\shortstack{Medium-\\resource}}
& pl & 0.995 & 0.966 & 0.977 & 0.916 & 0.227 & 0.704 \\
& nl & 0.996 & 0.970 & 0.945 & 0.922 & 0.524 & 0.183 \\
& ru & 0.998 & 0.938 & 0.956 & 0.946 & 0.220 & 0.429 \\
& hi & 0.994 & 0.980 & 0.941 & 0.916 & 0.639 & 0.061 \\
& ko & 0.988 & 0.957 & 0.979 & 0.711 & 0.335 & 0.616 \\
& ja & 0.992 & 0.965 & 0.954 & 0.820 & 0.439 & 0.272 \\
\midrule
\multirow{6}{*}{\shortstack{Low-\\resource}}
& bn & 0.993 & 0.960 & 0.943 & 0.795 & 0.365 & 0.323 \\
& fa & 0.993 & 0.978 & 0.944 & 0.861 & 0.600 & 0.250 \\
& vi & 0.994 & 0.972 & 0.955 & 0.783 & 0.285 & 0.191 \\
& iw & 0.996 & 0.977 & 0.899 & 0.899 & 0.361 & 0.130 \\
& uk & 0.998 & 0.958 & 0.949 & 0.904 & 0.278 & 0.177 \\
& ta & 0.994 & 0.969 & 0.961 & 0.837 & 0.269 & 0.172 \\
\bottomrule
\end{tabular}
\caption{\textbf{\textsc{Steam} \steam maintains strong AUC across all three text lengths}. Watermark strength (AUC and TPR@1\%) for short, medium, and long texts (bottom, middle, and top third by token count) using LLaMA-3.2 1B.}
\label{tab:text_length_llama}
\end{table*}
    
\clearpage

    \subsection{Ablation Study}
    \label{app:bo-ablation}
\begin{figure}[ht]
        \centering
        \includegraphics[width=1.0\linewidth]{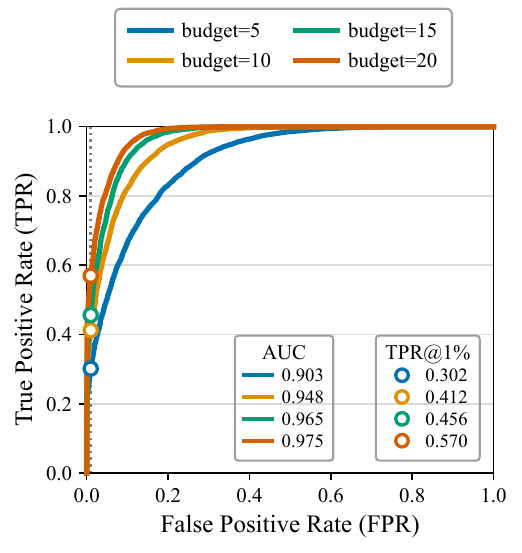}
        \caption{\textbf{\textsc{Steam} \steam achieves strong watermarking detection at the lowest optimisation budget.} ROC curves at budgets 5, 10, 15, and 20 on Aya-23 8B across 17 languages.}
        \label{fig:placeholder}
    \end{figure}

We ablate the Bayesian optimisation budget over values 5, 10, 15, and 20 on Aya-23 8B across 17 languages to characterise the accuracy-latency trade-off in \textsc{Steam} \steam. AUC improves monotonically from 0.903 to 0.975 and TPR@1\% from 0.302 to 0.570, with the largest gains concentrated between budgets 5 and 15  ($\Delta$AUC$=$~0.062, $\Delta$TPR@1\%~$=$~0.154, Figure~\ref{fig:placeholder}).

    \subsection{Adaptive Oracle Attacker}
    \label{app:oracle-attacker}

We evaluate an oracle adversary that selects the worst-case back-translation language across all 127 candidates.
In this setting, the adversary has full knowledge of the detection system and deliberately routes the watermarked text through the language that minimises \textsc{Steam}'s detection score.
Even when the adversary selects the single worst language (Uyghur), \textsc{Steam} \steam retains an AUC of 0.802 on Aya-23 8B with 500 samples (Table~\ref{tab:adaptive-oracle}), maintaining above-chance detection against a fully informed adversary.

    \begin{table}[H]
\centering
\small
\setlength{\tabcolsep}{2pt}
\begin{tabular}{l cc}
\toprule
\textbf{Setting} & \textbf{AUC} ($\uparrow$) & \textbf{TPR@1\%} ($\uparrow$) \\
\midrule
Mean over 126 languages          & 0.976 & 0.620 \\
Worst-10 oracle                  & 0.939 & 0.469 \\
Adaptive oracle (worst language) & 0.802 & 0.214 \\
\bottomrule
\end{tabular}
\caption{\textbf{\textsc{Steam} \steam retains detection capability even under worst-case oracle attacks.} AUC and TPR@1\% under three evaluation regimes: mean performance across 126 languages, a worst-10 oracle that averages the ten hardest languages, and a fully adaptive oracle that selects the single worst-case language.}
\label{tab:adaptive-oracle}
\end{table}

    \subsection{Paraphrase-then-translate Attack}
     \label{app:paraphrase-then-translate-attacker}
    
To challenge \textsc{Steam} \steam beyond single-step translation, we compose paraphrasing with back-translation: watermarked text is first paraphrased via GPT-4.1-nano, then back-translated on Gemma4-E4B across three languages. X-SIR collapses under this attack to a mean AUC and TPR@1\% of 0.492 and 0.000, respectively. In contrast, \textsc{Steam} \steam, retains an AUC above 0.967, demonstrating resilience against this stronger threat model (Table~\ref{tab:paraphrase-translate}).

    \begin{table*}[h!]
\centering
\small
\setlength{\tabcolsep}{3pt}
\begin{tabular}{lcccccccc}
\toprule
& \multicolumn{4}{c}{\textbf{AUC ($\uparrow$)}} & \multicolumn{4}{c}{\textbf{TPR@1\%} ($\uparrow$)} \\
\cmidrule(lr){2-5} \cmidrule(lr){6-9}
Lang. & KGW & X-KGW & X-SIR & STEAM \steam & KGW & X-KGW & X-SIR & STEAM \steam \\
\midrule
de & 0.699 & 0.787 & \red{0.539} & \textbf{0.975} & 0.148 & 0.252 & \red{0.000} & \textbf{0.494} \\
hi & 0.616 & 0.699 & \red{0.472} & \textbf{0.969} & 0.114 & 0.144 & \red{0.000} & \textbf{0.500} \\
iw & 0.628 & 0.645 & \red{0.466} & \textbf{0.967} & 0.104 & 0.122 & \red{0.000} & \textbf{0.458} \\
\bottomrule
\end{tabular}
\caption{\textbf{\textsc{Steam} \steam sustains robustness under a combined paraphrase-then-translate attack.} AUC and TPR@1\% for KGW, X-KGW, X-SIR, and \textsc{Steam} \steam on Gemma-4 4B with GPT-4.1-nano for paraphrasing. Red indicates values fall below the KGW baseline. Bolded is best.}
\label{tab:paraphrase-translate}
\end{table*}

\subsection{Calibration Ablation}
\label{app:calibration-ablation}
We also conduct an additional ablation study to assess the impact of per-language z-score normalization on STEAM’s performance. This component corrects the cross-lingual differences in the number of green tokens, which would otherwise increase the false positive rate. Removing normalization leads to only a modest AUC drop of 0.049. But it greatly reduces correct language identification, from 83.5\% to 38.6\% (Table~\ref{tab:kgw_llama_norm_summary}).
    \begin{table}[h!]
\centering
\small
\setlength{\tabcolsep}{1.7pt}
\begin{tabular}{lccc}
\toprule
\multirow{2}{*}{\makecell[b]{\textbf{Z-score}\\\textbf{normalization}}} & \multicolumn{2}{c}{\textbf{Watermark}} & \textbf{Language Pred.} \\
\cmidrule(lr){2-3} \cmidrule(lr){4-4}
 & AUC $\uparrow$ & TPR@1\% $\uparrow$ & Accuracy (\%) $\uparrow$ \\
\midrule
No                 & 0.902 {\tiny±0.11} & 0.34 {\tiny±0.13} & 38.6 {\tiny±16.6} \\
Yes (STEAM\:\steam) & 0.951 {\tiny±0.03} & 0.64 {\tiny±0.17} & 83.5 {\tiny±09.0} \\
\bottomrule
\end{tabular}
\caption{\textbf{Ablation study}. The z-score normalization in STEAM controls for differences in green token counts across languages. With normalization, STEAM selects the correct language and yields higher watermark strength. AUC, TPR@1\%, and language prediction accuracy (percentage of cases where the highest z-score corresponds to the back-translation to the original language), averaged over 17 languages with and without normalization.}
\label{tab:kgw_llama_norm_summary}
\end{table}

\clearpage

\section{Usage of AI Assistants}
\label{app:ai-assist}

For coding-related tasks, we relied on Claude 4.5 Sonnet and GitHub Copilot. We use GPT-5 and Claude for light editing (re-wording, grammar, proof-checking) to help writing the paper. For translation tasks in the experimental setting of \S\ref{subsec:steam_robustness}, we use DeepSeek-V3.2-Exp and GPT-4o-mini as the translation models.

\section{Artifacts}
\label{app:artifacts}

\subsection{Artifacts License}

All datasets, models, and code used in this work comply with their original licenses.

\begin{itemize}
    \item MUSE Dictionary\footnote{\url{https://github.com/facebookresearch/MUSE?tab=License-1-ov-file}} \cite{conneau2017word}: Released under the Creative Commons Attribution–NonCommercial 4.0 International (CC BY-NC 4.0) license. Use is restricted to non-commercial research and requires attribution to the original authors.
    \item Aya-23 8B\footnote{\url{https://huggingface.co/CohereLabs/aya-23-8B}}: Released under (CC BY-NC 4.0) license. 
    \item LLaMA-3.2 1B\footnote{\url{https://huggingface.co/meta-llama/Llama-3.2-1B/blob/main/LICENSE.txt}}: Released under the LLaMA 3.2 Community License Agreement. This license allows research and educational use but restricts commercial deployment without explicit permission from Meta.
    \item LLaMAX3 8B\footnote{\url{https://huggingface.co/LLaMAX/LLaMAX3-8B}}: Released under the MIT License, which permits reuse, modification, and redistribution for both commercial and non-commercial purposes, provided that attribution and the original license terms are preserved.
    \item DeepSeek-V3.2-Exp\footnote{\url{https://huggingface.co/deepseek-ai/DeepSeek-V3.2-Exp}}: Released under the MIT License.
    \item Gemma-4-E4B\footnote{\url{https://huggingface.co/google/gemma-4-E4B}}: Released under the Apache 2.0 License.
    \item mC4 Dataset\footnote{\url{https://huggingface.co/datasets/allenai/c4}} \cite{raffel2023exploringlimitstransferlearning}: Licensed under the Open Data Commons Attribution License (ODC-BY). This allows redistribution, reuse, and adaptation of the dataset, provided that appropriate credit is given.
    \item \texttt{deep\_translator}\footnote{\url{https://deep-translator.readthedocs.io/en/latest/README.html}} python package: Released under the MIT License.
    \item \texttt{openai}\footnote{\url{https://pypi.org/project/openai/}} python package: Released under the Apache License 2.0. This license permits use, modification, and redistribution for both commercial and non-commercial purposes
\end{itemize}
\subsection{Artifact Use Consistent With Intended Use}

All datasets and models were used in line with their intended research purposes and licences.
We used the mC4 dataset \citep{raffel2023exploringlimitstransferlearning} and open multilingual models (Aya-23-8B, LLaMA-3.2-1B, LLaMAX-8B) strictly for evaluation within academic settings. No data or model outputs were used for deployment or commercial applications. 
Our method \textsc{Steam} is released only for research use and is compatible with the original access conditions of all components. No personal data were processed. 







\end{document}